\documentclass[10pt,twocolumn,letterpaper]{article}

\usepackage{avss}
\usepackage{times}
\usepackage{epsfig}
\usepackage{graphicx}
\usepackage{amsmath}
\usepackage{amssymb}
\makeatletter
\@namedef{ver@everyshi.sty}{}
\makeatother
\usepackage{tikz,tikz-3dplot}
\usepackage{booktabs}
\usepackage{colortbl}
\usepackage{pgfplots}
\usepackage{comment}
\usepackage{adjustbox}
\usepackage[absolute,overlay]{textpos}

\usetikzlibrary{decorations.pathreplacing}
\usetikzlibrary{chains, positioning}
\usetikzlibrary{fadings,shapes.arrows,shadows} 
\usetikzlibrary{matrix,calc,positioning}
\usetikzlibrary{3d}

\definecolor{mblue}{RGB}{0,114,189}
\definecolor{mgreen}{RGB}{91,191,33}
\definecolor{morange}{RGB}{237,177,32}
\definecolor{mred}{RGB}{255,88,68}
\definecolor{mgray}{RGB}{145,140,136}
\definecolor{mbrown}{RGB}{215, 185, 101}
\definecolor{mredd}{RGB}{215, 109, 101}

\usepackage[pagebackref=true,breaklinks=true,letterpaper=true,colorlinks,bookmarks=false]{hyperref}

\avssfinalcopy 

\def\avssPaperID{****} 

\ifavssfinal\fi
\begin{document}


\title{ZSpeedL - Evaluating the Performance of Zero-Shot Learning Methods using Low-Power Devices}

\author{Cristiano Patrício and João C. Neves\\
NOVA-LINCS\\
University of Beira Interior, Portugal\\
{\tt\small cristiano.patricio@ubi.pt, jcneves@di.ubi.pt}
}

\maketitle
\thispagestyle{empty}


\begin{abstract}
The recognition of unseen objects from a semantic representation or textual description, usually denoted as zero-shot learning, is more prone to be used in real-world scenarios when compared to traditional object recognition. Nevertheless, no work has evaluated the feasibility of deploying zero-shot learning approaches in these scenarios, particularly when using low-power devices. 

In this paper, we provide the first benchmark on the inference time of zero-shot learning, comprising an evaluation of state-of-the-art approaches regarding their speed/accuracy trade-off. An analysis to the processing time of the different phases of the ZSL inference stage reveals that visual feature extraction is the major bottleneck in this paradigm, but, we show that lightweight networks can dramatically reduce the overall inference time without reducing the accuracy obtained by the \textit{de facto} ResNet101 architecture. Also, this benchmark evaluates how different ZSL approaches perform in low-power devices, and how the visual feature extraction phase could be optimized in this hardware. 

To foster the research and deployment of ZSL systems capable of operating in real-world scenarios, we release the evaluation framework used in this benchmark (\url{https://github.com/CristianoPatricio/zsl-methods}).

\end{abstract}

\section{Introduction}

Object recognition approaches are restricted to the identification of a finite number of classes observed during training. On the contrary, humans are capable of identifying unknown objects simply using a textual description. Based on this observation, zero-shot learning (ZSL) emerged as a more realistic alternative to traditional object recognition, where the goal is to build computational models capable of identifying objects solely with a semantic description of the class, and without samples in the training phase. 

In the last years, researchers have successfully exploited the advances on machine learning to boost the performance of this learning paradigm. At the same time, the effort put on creating standard evaluation protocols and benchmarks fostered the advances on the problem of ZSL. Thus, is not surprising that soon ZSL methods will be integrated into industrial solutions or end-user applications. Nevertheless, no work has evaluated the feasibility of deploying state-of-the-art ZSL methods in real-world scenarios, particularly the trade-off between accuracy and inference time.

\begin{figure*}[h!!]
\begin{center}
\resizebox{0.95\textwidth}{!}{
\begin{tikzpicture}
		
\def\a{.4}
\draw[yshift=0 mm] (1.65,6*\a+0.5)--++(90:\a)
--++(0:10.75) node[midway,above,xshift=4em]{\large Visual Feature Extraction}--++(-90:\a);

\draw[yshift=0 mm] (-0.75,2*\a-5.75)--++(90:\a)
--++(0:8.35) node[midway,above,xshift=2.5em]{\large Visual Feature Extraction}--++(-90:\a);
		
\draw[yshift=0 mm] (8.4,2*\a-5.75)--++(90:\a)
--++(0:5.5) node[midway,above,xshift=0em]{\large Feature Classification}--++(-90:\a);

		\node[rotate=90] at (-6.35,0) {\Large \textbf{Training Phase}};
		\node[rotate=90] at (-6.35,-6.75) {\Large \textbf{Inference Phase}};
		
		
		\newcommand{\networkLayer}[6]{
			\def\a{#1} 
			\def\b{0.02}
			\def\c{#2} 
			\def\t{#3} 
			\def\d{#4} 
			
			\draw[line width=0.9mm](\c+\t,0,\d) -- (\c+\t,\a,\d) -- (\t,\a,\d);                                                      
			\draw[line width=0.9mm](\t,0,\a+\d) -- (\c+\t,0,\a+\d) node[midway,below] {#6} -- (\c+\t,\a,\a+\d) -- (\t,\a,\a+\d) -- (\t,0,\a+\d); 
			\draw[line width=0.9mm](\c+\t,0,\d) -- (\c+\t,0,\a+\d);
			\draw[line width=0.9mm](\c+\t,\a,\d) -- (\c+\t,\a,\a+\d);
			\draw[line width=0.9mm](\t,\a,\d) -- (\t,\a,\a+\d);

			\filldraw[#5] (\t+\b,\b,\a+\d) -- (\c+\t-\b,\b,\a+\d) -- (\c+\t-\b,\a-\b,\a+\d) -- (\t+\b,\a-\b,\a+\d) -- (\t+\b,\b,\a+\d); 
			\filldraw[#5] (\t+\b,\a,\a-\b+\d) -- (\c+\t-\b,\a,\a-\b+\d) -- (\c+\t-\b,\a,\b+\d) -- (\t+\b,\a,\b+\d);

			\filldraw[#5] (\c+\t,\b+0.05,\a-\b+\d-0.1) -- (\c+\t,\b+0.05,\b+\d+0.09) -- (\c+\t,\a-\b-0.05,\b+\d+0.1) -- (\c+\t,\a-\b-0.05,\a-\b+\d-0.1); 
		}



\foreach \x in {1,...,5} {
    \begin{scope}[xshift=-3.95cm,yshift=0.85cm,cm={0.7,0.5,0,1,(0.65*\x,0)}]
      \node[transform shape, draw, black, ultra thick, inner sep=0.05mm] at(0,0) {
        \includegraphics[width=4cm]{figs/img\x.jpg}
      };
    \end{scope}
  }

\def\nx{1.2}
\networkLayer{3.5}{0.2}{2+\nx}{0.0}{color=mblue!60}{}
\networkLayer{3.5}{0.2}{2.4+\nx}{0.0}{color=morange!60}{}

\networkLayer{3}{0.2}{3.5+\nx}{0.0}{color=mblue!60}{} 
\networkLayer{3}{0.2}{3.9+\nx}{0.0}{color=morange!60}{} 

\networkLayer{2.25}{0.2}{4.9+\nx}{0.0}{color=mblue!60}{} 
\networkLayer{2.25}{0.2}{5.3+\nx}{0.0}{color=morange!60}{}

\networkLayer{1.5}{0.2}{6.2+\nx}{0.0}{color=mblue!60}{} 
\networkLayer{1.5}{0.2}{6.55+\nx}{0.0}{color=morange!60}{}


\path [draw,ultra thick,->] (1.9+\nx,1)-- (2.3+\nx,1);
\path [draw,ultra thick,->] (3.5+\nx,0.7)-- (3.9+\nx,0.7);
\path [draw,ultra thick,->] (5.1+\nx,0.5)-- (5.5+\nx,0.5);

\tikzset{arraystyle/.style={draw, font=\fontsize{10}{9}, minimum width=1.35em, minimum height=1em, outer sep=0pt},}

\begin{scope}[xshift=9.5cm, yshift=1.2cm, rotate around x=25]
\matrix (ma) at (0,0) [matrix of math nodes, ampersand replacement=\&, nodes={arraystyle, anchor=center,fill=white}, column sep=-\pgflinewidth]
{
1.3   \\
0.9   \\
1.3   \\
\cdots  \\
2.0   \\
1.2   \\
0.3 \\
};

\matrix (mb) at ($(ma.south west)+(1.35,2)$)  [matrix of math nodes, ampersand replacement=\&, nodes={arraystyle, anchor=center,fill=white}, column sep=-\pgflinewidth]
{
0.2 \\
2.5 \\
1.1 \\
\cdots \\
1.6 \\
3.4 \\
0.8 \\
};

\matrix (mc) at ($(mb.south west)+(1.35,2)$)  [matrix of math nodes, ampersand replacement=\&, nodes={arraystyle, anchor=center,fill=white}, column sep=-\pgflinewidth]
{
0.2 \\
2.5 \\
1.1 \\
\cdots \\
1.6 \\
3.4 \\
0.8 \\
};

\matrix (md) at ($(mc.south west)+(1.35,2)$)  [matrix of math nodes, ampersand replacement=\&, nodes={arraystyle, anchor=center,fill=white}, column sep=-\pgflinewidth]
{
0.2 \\
2.5 \\
1.1 \\
\cdots \\
1.6 \\
3.4 \\
0.8 \\
};

\matrix (me) at ($(md.south west)+(1.35,2)$)  [matrix of math nodes, ampersand replacement=\&, nodes={arraystyle, anchor=center,fill=white}, column sep=-\pgflinewidth]
{
0.2 \\
2.5 \\
1.1 \\
\cdots \\
1.6 \\
3.4 \\
0.8 \\
};

\end{scope}

\def\dx{1.75}
\def\dy{-1.5}
\node[anchor=west] at (0-\dx,0+\dy) {stripes:};
\node[anchor=west] at (0-\dx,-0.4+\dy) {big:};
\node[anchor=west] at (0-\dx,-0.8+\dy) {swim:};
\node[anchor=west] at (0-\dx,-1.2+\dy) {hunter:};
\node[anchor=west] at (0-\dx,-1.6+\dy) {tail:};

\node[anchor=east] at (2.8-\dx,0+\dy) {no};
\node[anchor=east] at (2.8-\dx,-0.4+\dy) {yes};
\node[anchor=east] at (2.8-\dx,-0.8+\dy) {no};
\node[anchor=east] at (2.8-\dx,-1.2+\dy) {no};
\node[anchor=east] at (2.8-\dx,-1.6+\dy) {yes};

\draw [->, thick] (12.25,1.5) to [out=0,in=180] (13.75,1.5);
\draw [->, thick] (14,1.5) to [out=0,in=120] (15.6,-0.25);
\draw [->, thick] (14,-2.5) to [out=340,in=210] (15.65-0.25,-1.1);
\draw [->, thick] (1.25,-2.5) to [out=0,in=180] (13.45,-2.5);

\draw [fill=gray!50, draw=none, ultra thick] (13.35,2) rectangle (14,-3) node[pos=.5, rotate=90] {\textbf{ZSL Method}};

\begin{axis}[anchor = north west,xmin=-0,xmax=0.99,
  name={ax2}, 
  domain=0:1, 
  samples=1,
  ymin=0,
  ymax=0.99,
  axis x line = middle, 
  axis y line = middle, 
  xlabel={},
  ylabel={},
  xticklabels={,,},
  yticklabels={,,},
  height=4.75cm, 
  width=4.75cm,
  enlargelimits=false, 
  clip=false, 
  grid = minor,
  at = {(14.8 cm,1.5 cm)}
  ] 

\node[label={180:{}},circle,fill,inner sep=1.5pt] at (axis cs:0.2,0.2) {};
\node[label={180:{}},circle,fill,inner sep=1.5pt] at (axis cs:0.25,0.4) {};

\end{axis}


\draw [-] (-5.7,-3.5) to [out=0,in=180] (19.25,-3.5);

\def\dxx{0.9}
\def\dyy{-4.7}
\node (img2) at (-4.0+\dxx,-2.5+\dyy)
{\includegraphics[width=3.85cm]{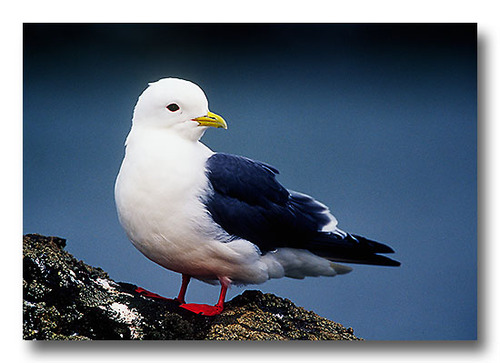}};

\node[] at (0.14+\dxx,6.2+\dyy) {\LARGE $\boldsymbol{X^{s}}$};
\node at (-4.0+\dxx,-0.75+\dyy) {\LARGE $\boldsymbol{X^{u}}$};
\node at (15.4+\dxx,0.0+\dyy) {\LARGE $\boldsymbol{C^{u}}$};
\node at (18+\dxx,-0.65+\dyy) {\LARGE $\boldsymbol{Y^{u}}$};

\begin{scope}[yshift=-8cm, xshift=-1cm]
\networkLayer{3.5}{0.2}{2}{0.0}{color=mblue!60}{}
\networkLayer{3.5}{0.2}{2.4}{0.0}{color=morange!60}{}

\networkLayer{3}{0.2}{3.5}{0.0}{color=mblue!60}{} 
\networkLayer{3}{0.2}{3.9}{0.0}{color=morange!60}{} 

\networkLayer{2.25}{0.2}{4.9}{0.0}{color=mblue!60}{} 
\networkLayer{2.25}{0.2}{5.3}{0.0}{color=morange!60}{}

\networkLayer{1.5}{0.2}{6.2}{0.0}{color=mblue!60}{} 
\networkLayer{1.5}{0.2}{6.55}{0.0}{color=morange!60}{} 

\path [draw,ultra thick,->] (1.9,1)-- (2.3,1);
\path [draw,ultra thick,->] (3.5,0.7)-- (3.9,0.7);
\path [draw,ultra thick,->] (5.1,0.5)-- (5.5,0.5);

\end{scope}

\matrix at (6.25+\dxx,-2.5+\dyy) (A) [matrix of math nodes, ampersand replacement=\&, column 1/.style={minimum width=3em,nodes={minimum width=200em,arraystyle}}]
{2.1 \\ 0.5 \\ 0.1 \\ \cdots \\ 1.9 \\ 2.3\\ 0.3\\};

\matrix at (15.25+\dxx,-2.25+\dyy) [matrix of nodes, align=center,nodes={draw},text width=2em,text depth=.5ex,text height=1.5ex,row 3/.style={nodes={fill=mbrown!35}}, ampersand replacement=\&, font=\small] {
      $Att_1$ \& $Att_2$ \& $\cdots$ \& $Att_n$ \\
      $yes$ \& $no$ \& $\cdots$ \& $yes$ \\
      $yes$ \& $no$ \& $\cdots$ \& $yes$ \\
      $yes$ \& $no$ \& $\cdots$ \& $yes$ \\
      $\cdots$ \& $\cdots$ \& $\cdots$ \& $\cdots$ \\
      $yes$ \& $no$ \& $\cdots$ \& $yes$ \\
    };

\matrix at (17.9+\dxx,-2.525+\dyy) [matrix of nodes, align=left, nodes={}, text width=2.85em,text depth=.5ex,text height=1.5ex, font=\small] {
      $zebra$ \\
      $bird$ \\
      $horse$ \\
      $\cdots$ \\
      $tiger$ \\
    };

\draw [->] (6.6+\dxx,-2.5+\dyy) to [out=0,in=180] (7.5+\dxx,-2.5+\dyy);
\draw [->] (8.15+\dxx,-2.5+\dyy) to [out=-45,in=225] (9.7+\dxx,-2.75+\dyy);

\draw [fill=gray!50, draw=none, ultra thick] (7.5+\dxx,-4+\dyy) rectangle (8.15+\dxx,-1+\dyy) node[pos=.5, rotate=90] {\textbf{ZSL Method}};

\draw [fill=gray!50, draw=none, ultra thick] (12+\dxx,-4+\dyy) rectangle (12.65+\dxx,-1+\dyy) node[pos=.5, rotate=90] {\textbf{ZSL Method}};

\draw [->] (13.3+\dxx,-1.4+\dyy) to [out=180,in=0] (12.7+\dxx,-1.4+\dyy);
\draw [->] (13.3+\dxx,-1.95+\dyy) to [out=180,in=0] (12.7+\dxx,-1.95+\dyy);
\draw [->] (13.3+\dxx,-2.5+\dyy) to [out=180,in=0] (12.7+\dxx,-2.5+\dyy);
\draw [->] (13.3+\dxx,-3.6+\dyy) to [out=180,in=0] (12.7+\dxx,-3.6+\dyy);

\draw [->] (12+\dxx,-1.4+\dyy) to [out=210,in=-30] (11+\dxx,-1.2+\dyy);
\draw [->] (12+\dxx,-1.95+\dyy) to [out=210,in=-30] (9.75+\dxx,-2+\dyy);
\draw [->] (12+\dxx,-2.5+\dyy) to [out=180,in=45] (11+\dxx,-3+\dyy);
\draw [->] (12+\dxx,-3.6+\dyy) to [out=180,in=0] (9.8+\dxx,-4+\dyy);

\begin{axis}[anchor = north west,xmin=-0,xmax=0.99,
  name={ax2}, 
  domain=0:1, 
  samples=1,
  ymin=0,
  ymax=0.99,
  axis x line = middle, 
  axis y line = middle, 
  xlabel={},
  ylabel={},
  xticklabels={,,},
  yticklabels={,,},
  height=4.75cm, 
  width=4.75cm,
  enlargelimits=false, 
  clip=false, 
  grid = minor,
  at = {(9.9 cm,-6.15 cm)}
  ] 

\node[mark size=2.5pt,label={180:{}},inner sep=1.5pt] at (axis cs:0.585,1.115) {\pgfuseplotmark{triangle*}};
\node[mark size=2.5pt,label={180:{}},inner sep=1.5pt] at (axis cs:0.19,0.85) {\pgfuseplotmark{triangle*}};
\node[mark size=2.5pt,label={180:{}},inner sep=1.5pt] at (axis cs:0.6,0.5) {\pgfuseplotmark{triangle*}};
\node[mark size=2.5pt,label={180:{}},inner sep=1.5pt] at (axis cs:0.21,0.2) {\pgfuseplotmark{triangle*}};
\node[mark size=2.5pt,label={180:{}},inner sep=1.5pt] at (axis cs:0.21,0.2) {\pgfuseplotmark{triangle*}};
\node[mark size=2.5pt,circle, fill,label={180:{}},inner sep=1.5pt] at (axis cs:0.26,0.64) {};


\end{axis}

\end{tikzpicture}
}
\end{center}
\caption{\textbf{Processing chain of zero-shot learning.} Visual feature extraction allows representing images through a compact descriptor. These visual features and its corresponding semantic embedding are fed, during the training phase, to ZSL methods for learning a function that reduces the distance between the two representations. In the inference phase, the visual features of the query image are transformed by the ZSL method, and the class of the images is determined by the closest semantic embedding of the test set.\label{fig1}}
\end{figure*}


In this paper, we introduce the first benchmark on the accuracy and inference time of ZSL methods, intended to study the feasibility of deploying these methods in real-world scenarios using low-power computational devices. This benchmark evaluates the impact of the two major phases of the processing chain of zero-shot learning, illustrated in Figure~\ref{fig1}. The experimental results show that visual feature extraction is the bottleneck in the processing chain of these approaches, and thus this work is also concerned on assessing the impact that lighter CNN architectures have in the overall speed and accuracy of ZSL approaches. Also, our experiments show that is possible to reduce the size of the architecture used without significantly impacting the ZSL method's accuracy. This benchmark also reports the number of frames per second analyzed when using different low-power hardware devices, providing insights about the workability of these approaches in real-world scenarios. Finally, to foster the research on zero-shot learning and the development of ZSL-based solutions, we release the source-code for reproducing all the experiments performed, as well as the instructions for optimizing and deploying ZSL methods in low-power devices.

\vspace{0.01cm}

Accordingly, the major contributions of this paper are the following: (1) An extensive benchmark across four widely known ZSL datasets regarding the accuracy and processing time of state-of-the-art ZSL methods using visual features obtained from different CNN architectures. (2) A comparative analysis of the processing time of ZSL algorithms using lightweight architectures when run in different low-power devices. (3) An open-source evaluation framework for analyzing the accuracy/speed trade-off in the problem of ZSL.

The remainder of this paper is organized as follows: section 2 summarizes the most relevant works in the scope of our work. Section 3 provides the details of the experiments performed. In section 4 we discuss the obtained results and the conclusions are given in section 5.

\section{Related Work}


\subsection{ZSL Problem\label{sec:zsl_problem}}

Zero-shot learning regards the problem of identifying classes not observed during the training phase. Formally, given a set of images $X = X^{s} \cup X^{u}$, and the corresponding labels of seen (s) and unseen (u) data, $Y = Y^s \cup Y^u$, the goal of ZSL is inferring $X^{u} \to Y^{u}$, without having access to $Y^{u}$ during training. The only ancillary information provided is the semantic representation $C = C^s \cup C^u$ describing each class, which should be used as a proxy to map the visual data to the unseen classes. For inferring this mapping, most works follow one of two major strategies: 1) learning a projection function capable of relating both encodings in a common low-dimensional space; 2) generating the visual features from the semantic embedding for unseen classes and train a standard classifier to distinguish between classes. The work of Lampert \etal~\cite{Lampert_2009_CVPR} was precursor of projection-based methods where a binary classifier was trained to predict the attributes of an image given its visual features. In the test phase, an image was classified by measuring the probability of the predicted attributes match the semantic representation of unseen classes. Later, the idea of learning a linear compatibility matrix between visual features and semantic information was explored in ~\cite{Akata_CVPR2013}. The use of inner product similarity allowed to devise a simple system of linear operations capable of determining the similarity between the image features and unseen classes. This idea was explored in several works where different loss functions~\cite{Frome_NIPS20213, Akata_CVPR2015} and regularization terms were proposed~\cite{Romera_Paredes_ICML2015}. Also, several works improved the original idea by using non-linear compatibility functions~\cite{Socher_NIPS2013, Xian_CVPR2016}. A different strategy was projecting both visual features and semantic embeddings into a common sub-space. Zhang and Saligrama~\cite{Zhang_CVPR2016} proposed the learning of a class-independent function to map both representations to a latent representations. Changpinyo \etal~\cite{Changpinyo_CVPR2016} relied on the notion of a weighted graph to align the semantic space with the model space, composed of classifiers for visual recognition, such that the coordinates of the model space are a projection of the graph vertices from the semantic space. Zhang \etal~\cite{Zhang_CVPR2017} proposed the first deep end-to-end ZSL model by training simultaneously the CNN used for visual feature extraction and a set of fully connected layers for transforming the semantic representation to the visual space. They found that this model coupled with the least square loss between the two embedded vectors is less prone to suffer from the hubness problem. 

In spite of the advances obtained with projection methods, they failed to achieve reasonable results in the generalized setting of zero-shot learning, where both seen and unseen classes are available in the test phase. This is explained by the bias of the mapping function towards seen classes, and to address this issue,  Xian \etal.~\cite{Xian_CVPR2018} suggested the idea of generating the visual features of unseen classes conditioned on semantic information, and rely on a supervised classifier to distinguish between the target classes using the features extracted from the query image.
Following this idea, several approaches have been proposed using different models for generating features~\cite{Xian_CVPR2018}, the use of cycle-consistency loss~\cite{Xian_CVPR2019}, or the use of feedback loops for refining generated features~\cite{Narayan_2020}.

\subsection{Visual Feature Extraction}

ZSL methods rely on visual feature extraction to represent the images in a compact manner. While initial ZSL approaches used hand-crafted techniques for feature extraction, the use of Convolutional Neural Networks (CNN) has become the standard procedure for obtaining a compact description of a test image. Considering this, the research on ZSL has been primarily focused on devising strategies to learn the mapping from visual representations to semantic embeddings. As described in section~\ref{sec:zsl_problem}, these strategies can be broadly divided into two major groups: 1) projection-based methods; 2) generative methods. The computational burden of each family is analyzed in Section~\ref{resuts_rq1}.

\subsection{ZSL Computational Cost}

Most contributions on ZSL disregard the visual feature extraction phase by using precomputed features from standard CNN architectures. The reduced size of these features significantly decreases the processing time of the feature classification in the ZSL processing chain, even when the classification algorithm has a high temporal complexity. For this reason, most works have focused solely on reporting accuracy, and few works have analyzed inference time of the proposed models. In~\cite{Wang_Neurocomputing2021} and~\cite{Liu_NN2020}, the authors perform an analysis on the complexity of the learning strategy. Ji \etal~\cite{Ji_IS2017} report the processing time during both the training and inference phase, and provide the complexity of the proposed algorithm. Pan \etal~\cite{Pan_Neuralcomputing2020} evaluate the inference time of the proposed method both on CPU and GPU, as well as the time required by competing approaches. In other work~\cite{Li_PRL2020}, the authors report the processing time of the proposed method when using different datasets, as they noticed differences according to the number of unseen classes. 

Nevertheless, no work has specifically evaluated the inference time of the overall processing chain of ZSL, neither the impact of using different architectures for the visual feature extraction phase. Also, the evaluation on low-power devices has not been considered yet. To the best of our knowledge, our work is the first benchmark on ZSL inference time in low-power devices, providing a comparative analysis of how different CNN architectures impact the speed/accuracy trade-off of these approaches.

\section{Evaluation Methodology}

This section details the experimental protocol, including a description of the testbed state-of-the-art ZSL methods, the statistics of used datasets, and the adopted evaluation protocols.

\subsection{Methods}

We have selected six state-of-the-art ZSL methods, including ESZSL \cite{Romera_Paredes_ICML2015}, SAE \cite{Koridov_CVPR2017}, DEM \cite{Zhang_CVPR2017}, f-CLSWGAN \cite{Xian_CVPR2018}, TF-VAEGAN \cite{Narayan_2020}, and CE-GSZL \cite{Han_2021}. The selected approaches cover the two major strategies in ZSL: (1) projection-based methods (ESZSL \cite{Romera_Paredes_ICML2015}, SAE \cite{Koridov_CVPR2017}, and DEM \cite{Zhang_CVPR2017}), and (2) generative methods (f-CLSWGAN \cite{Xian_CVPR2018}, TF-VAEGAN \cite{Narayan_2020}, and CE-GZSL \cite{Han_2021}). 

\textbf{DAP}~\cite{Lampert_2009_CVPR} was a seminal work in the topic of ZSL, where a binary classifier was trained to predict the attributes of an image given its visual features. In the inference phase, an image is classified by measuring the probability of the predicted attributes match the semantic representation of unseen classes. Regarding its computational cost, this approach depends mainly on complexity of the classifier used to estimate the attributes of an image given its visual representation and the number of test classes. 

\textbf{ESZSL}~\cite{Romera_Paredes_ICML2015} is a representative approach of the use of inner product similarity for determining the similarity between the image features and image classes. Similarity scores for test classes are inferred through a set of linear transformations, encoded by matrix multiplication operations. The reduced complexity of these transformations, and the possibility of parallelizing matrix multiplication, ensures a short inference time.

\textbf{SAE}~\cite{Koridov_CVPR2017} treats ZSL as an encoding-decoding problem by using an auto-encoder (AE) to transform the visual features to the semantic space and subsequently recover the same features from the class semantic representation. This strategy significantly outperformed state-of-the-art, and allowed to classify an image either by inferring its semantic representation from the visual features, or transforming the semantic representation of target classes to the visual feature space and then use k-NN in the feature space. The use of a linear AE allows a very fast attribute estimation. However, the overall inference time is mainly dependent on the size of unseen classes due to the use of k-NN.

\textbf{DEM}~\cite{Zhang_CVPR2017} was one of the first end-to-end ZSL models where the CNN and the feature projection function  were jointly optimized. This strategy significantly increases training time, but the classification inference phase only depends on the size and number of fully connected layers.  

\textbf{f-CLSWGAN}~\cite{Xian_CVPR2018} introduced the disruptive idea of casting ZSL as a traditional supervised classification problem, by training a GAN for generating the visual features of any class from their semantic representation, and use a standard supervised classifier to distinguish between the unseen classes. During the test phase, inference time is only dependent on the classifier used (softmax classifier in the original work).

\textbf{TF-VAEGAN}~\cite{Narayan_2020} constitutes an improvement over traditional generative approaches, since it increases the semantic consistency of generated features by enforcing the decoding of generated features to be similar to original semantic embeddings. This semantic embedding decoder is also used during the inference stage for providing the feature classifier with both extracted features and their decoded semantic embeddings, which slightly increases the inference time over f-CLSWGAN.

\textbf{CE-ZSL}~\cite{Han_2021} is a hybrid approach combining a projection model and a generative model. The projection model is used to estimate the semantic embeddings of generated visual features, in order to compare these embeddings with the original ones using comparator network trained using contrastive learning. Contrary to other generative approaches, CE-ZSL performs the classification in the semantic space, where a softmax classifier is trained using both seen semantic information  and unseen synthetic embeddings. For this reason, as in the case of f-CLSWGAN, inference time depends solely on the classifier used.

\subsection{Datasets}
\label{datasets}

The evaluation of the six state-of-the-art ZSL methods considered in this study is carried out on the four most popular ZSL datasets, namely, \textit{Animals with Attributes 2} (AWA2) \cite{Xian_2018_TPAMI}, \textit{Caltech-UCSD-Birds} (CUB-200-2011) \cite{WahCUB_200_2011}, \textit{SUN Attributes} \cite{SUN} (SUN), and \textit{Attribute Pascal and Yahoo} (aPascal-aYahoo) \cite{Farhadi_2009_CVPR}. The datasets statistics are presented in Table \ref{tab:datasets}.

\begin{table}[h]
\caption{Statistics of the datasets used in this benchmark.\label{tab:datasets}}
	\centering
	    \scalebox{0.75}{
		\begin{tabular}{|c|c|c|c|}
		\hline
			\textbf{Dataset} & \textbf{No. Classes} & \textbf{No. Instances} & \textbf{No. Attributes} \cr
			\hline
			CUB-200-2011 \cite{WahCUB_200_2011} & 200 & 11788 & 312 \cr
			\hline
			SUN Attributes \cite{SUN} & 717 & 14340 & 102 \cr
			\hline
			AWA2 \cite{Xian_2018_TPAMI} & 50 & 37322 & 85 \cr
			\hline
			aPascal-aYahoo \cite{Farhadi_2009_CVPR} & 32 & 15339 & 64 \cr
			\hline
		\end{tabular}
	}
\end{table}


The visual features for all the images in the dataset are extracted using the top-layer pooling units of the ResNet-101, MobileNet, MobileNetV2, Xception, and EfficientNetB7 pre-trained on ImageNet-1K without fine-tuning. Moreover, we adopt the Proposed Split (PS)~\cite{Xian_2018_TPAMI} to ensure that test classes are disjoint from the ones used to train the CNN model. For semantic embeddings, we use the class-level attributes provided by~\cite{Xian_2018_TPAMI} for AWA2 (85-dim), SUN (102-dim), CUB (312-dim), and APY (64-dim). We conduct all experiments under the inductive setting, in which only labeled instances of seen classes are considered at the training phase~\cite{Xian_2018_TPAMI}.

\subsection{Hardware}

Experiments were performed in a desktop computer and two small low-power devices, namely a Raspberry Pi 4 Model B and a Jetson Nano Developer Kit. The hardware specifications are given in table \ref{tab:hw_specs}.

\begin{table}[h]
\centering
\caption{Hardware Specification.\label{tab:hw_specs}}

    \resizebox{0.48\textwidth}{!}{
    \begin{tabular}{|c|c|c|c|}
        \hline
                                      & \textbf{Desktop}                                                                & \textbf{Raspberry Pi 4B} & \textbf{Jetson Nano Dev Kit}\\ 
        \hline
        \emph{\textbf{CPU}} & \begin{tabular}[c]{@{}c@{}}Intel® Core™ i7-10700U\\ CPU @ 2.90GHz × 16\end{tabular} & \begin{tabular}[c]{@{}c@{}} Broadcom BCM2711 \\ Quad core Cortex-A72 (ARM v8)\\ 64-bit SoC @ 1.5GHz\end{tabular} & Quad-core ARM A57 @ 1.43 GHz\\ 
        \hline
        \emph{\textbf{GPU}}     & -    & - & 128-core Maxwell\\ 
        \hline
        \emph{\textbf{RAM}} & 32GB & 4GB & 4 GB\\ 
        \hline
        \emph{\textbf{Storage}} & 1TB SSD                                                                & 64GB microSD & 64GB microSD\\ 
        \hline
        \emph{\textbf{OS}} & Pop\_OS! 20.10 64-bit                                                                & Raspbian & Ubuntu 18.04.5 LTS\\ 
        \hline
    \end{tabular}
    }
\end{table}

\subsection{Evaluation Protocols}

The performance of ZSL methods in the restricted and generalized settings is measured using the standard evaluation protocol proposed in~\cite{Xian_2018_TPAMI}. The multi-way classification accuracy (MCA) is adopted to assess the average per-class accuracy in the restricted setting. In the generalized setting, the average per-class classification accuracy is determined on training ($Y^{s}$) and test ($Y^{u}$) classes, and the harmonic mean is obtained by $H = \frac{2 * acc_{Y^{s}} * acc_{Y^{u}}}{acc_{Y^{s}} + acc_{Y^{u}}}$, where $acc_{Y^{s}}$ and $acc_{Y^{u}}$ denotes the accuracy of seen and unseen classes, respectively.

\section{Results}


\subsection{ZSL Methods: Inference Time\label{resuts_rq1}}

This section analyzes the time consumed by different ZSL methods for classifying a single test image, disregarding the visual feature extraction phase. This analysis intends to evaluate the impact of the different feature classification strategies in the overall inference stage of ZSL methods. Moreover, the impact of the size of visual features and the use of low-power devices in the processing time is also studied. The results obtained are reported in Table~\ref{tab:zsl_time}.

\begin{table*}[h]
    \caption{Processing time (in milliseconds) of feature classification on CPU, Raspberry Pi 4B (R-PI 4B) and Jetson Nano. \label{tab:zsl_time}}
    \centering
    \resizebox{0.98\textwidth}{!}{
    \begin{tabular}{|l|c|c|c|c|c|c|c|c|c|c|c|c|}
        \hline
        \multicolumn{1}{|c|}{} & \multicolumn{12}{|c|}{Visual Features Dimension} \cr
        \hline
        \multicolumn{1}{|c|}{} & \multicolumn{3}{|c|}{\textbf{512}} & \multicolumn{3}{|c|}{\textbf{1,024}} & \multicolumn{3}{|c|}{\textbf{2,048}} & \multicolumn{3}{|c|}{\textbf{4,032}} \cr
        \hline
        \textbf{Method} & Desktop & R-PI 4B & Jetson Nano & Desktop & R-PI 4B & Jetson Nano & Desktop & R-PI 4B & Jetson Nano & Desktop & R-PI 4B & Jetson Nano \cr
        \hline
        DAP & $0.74 {\scriptstyle \pm 0.00}$ & $18.72 {\scriptstyle\pm 0.45}$ & $5.11 {\scriptstyle\pm 0.05}$ 
            & $0.43 {\scriptstyle\pm 0.00}$ & $19.05 {\scriptstyle\pm 0.31}$ & $5.24 {\scriptstyle\pm 0.15}$
            & $0.45 {\scriptstyle\pm 0.01}$ & $18.82 {\scriptstyle\pm 0.32}$ & $5.20 {\scriptstyle\pm 0.10}$
            & $0.44 {\scriptstyle\pm 0.02}$ & $18.61 {\scriptstyle\pm 0.35}$ & $5.28 {\scriptstyle\pm 0.11}$ \cr
        \hline
        IAP & $0.74 {\scriptstyle\pm 0.00}$ & $18.61 {\scriptstyle\pm 0.20}$ & $5.11 {\scriptstyle\pm 0.03}$
            & $0.44 {\scriptstyle\pm 0.00}$ & $18.85 {\scriptstyle\pm 0.07}$ & $5.28 {\scriptstyle\pm 0.13}$
            & $0.45 {\scriptstyle\pm 0.02}$ & $18.70 {\scriptstyle\pm 0.08}$ & $5.24 {\scriptstyle\pm 0.09}$
            & $0.43 {\scriptstyle\pm 0.00}$ & $18.49 {\scriptstyle\pm 0.06}$ & $5.19 {\scriptstyle\pm 0.11}$ \cr
        \hline
        SAE & $0.12 {\scriptstyle\pm 0.00}$ & $1.11 {\scriptstyle\pm 0.13}$ & $1.21 {\scriptstyle\pm 0.05}$
            & $0.12 {\scriptstyle\pm 0.00}$ & $1.35 {\scriptstyle\pm 0.09}$ & $1.57 {\scriptstyle\pm 0.02}$
            & $0.13 {\scriptstyle\pm 0.00}$ & $1.66 {\scriptstyle\pm 0.07}$ & $1.88 {\scriptstyle\pm 0.05}$
            & $0.13 {\scriptstyle\pm 0.00}$ & $2.29 {\scriptstyle\pm 0.07}$ & $2.42 {\scriptstyle\pm 0.04}$ \cr
        \hline
        ESZSL & $0.03 {\scriptstyle\pm 0.00}$ & $0.81 {\scriptstyle\pm 0.09}$ & $0.49 {\scriptstyle\pm 0.05}$ 
              & $0.03 {\scriptstyle\pm 0.00}$ & $1.08 {\scriptstyle\pm 0.10}$ & $0.75 {\scriptstyle\pm 0.06}$
              & $0.04 {\scriptstyle\pm 0.00}$ & $1.40 {\scriptstyle\pm 0.15}$ & $0.97 {\scriptstyle\pm 0.07}$
              & $0.06 {\scriptstyle\pm 0.00}$ & $1.89 {\scriptstyle\pm 0.05}$ & $1.37 {\scriptstyle\pm 0.02}$ \cr
        \hline
        DEM & $0.83 {\scriptstyle\pm 0.05}$ & $4.15 {\scriptstyle\pm 0.03}$ & $9.43 {\scriptstyle\pm 0.49}$  
            & $1.54 {\scriptstyle\pm 0.12}$ & $7.35 {\scriptstyle\pm 0.12}$ & $13.33 {\scriptstyle\pm 0.32}$
            & $3.11 {\scriptstyle\pm 0.14}$ & $26.74 {\scriptstyle\pm 0.80}$ & $24.28 {\scriptstyle\pm 1.47}$
            & $6.25 {\scriptstyle\pm 0.23}$ & $44.69 {\scriptstyle\pm 1.28}$ &  $40.29 {\scriptstyle\pm 2.94}$ \cr
        \hline
        f-CLSWGAN & $0.71 {\scriptstyle\pm 0.06}$ & $3.95 {\scriptstyle\pm 0.07}$ & $2.02 {\scriptstyle\pm 0.12}$ 
                  & $0.83 {\scriptstyle\pm 0.09}$ & $4.40 {\scriptstyle\pm 0.09}$ & $1.98 {\scriptstyle\pm 0.09}$
                  & $0.96 {\scriptstyle\pm 0.10}$ & $5.72 {\scriptstyle\pm 0.12}$ & $1.79 {\scriptstyle\pm 0.05}$
                  & $1.26 {\scriptstyle\pm 0.11}$ & $7.03 {\scriptstyle\pm 0.14}$ & $2.55 {\scriptstyle\pm 0.13}$ \cr
        \hline
        TF-VAEGAN & $1.40 {\scriptstyle\pm 0.03}$ & $17.89 {\scriptstyle\pm 0.87}$ & $10.49 {\scriptstyle\pm 1.90}$ 
                  & $1.97 {\scriptstyle\pm 0.05}$ & $26.83 {\scriptstyle\pm 1.65}$ & $12.02 {\scriptstyle\pm 0.39}$
                  & $3.08 {\scriptstyle\pm 0.06}$ & $44.95 {\scriptstyle\pm 2.53}$ & $10.23 {\scriptstyle\pm 0.76}$
                  & $5.35 {\scriptstyle\pm 0.10}$ & $77.28 {\scriptstyle\pm 3.23}$ & $19.42 {\scriptstyle\pm 4.76}$ \cr
        \hline
        CE-GZSL & $0.75 {\scriptstyle\pm 0.09}$ & $4.08 {\scriptstyle\pm 0.22}$ & $2.70 {\scriptstyle\pm 0.09}$ 
                  & $0.82 {\scriptstyle\pm 0.09}$ & $4.38 {\scriptstyle\pm 0.24}$ & $1.76 {\scriptstyle\pm 0.16}$
                  & $0.95 {\scriptstyle\pm 0.06}$ & $5.33 {\scriptstyle\pm 0.13}$ & $1.71 {\scriptstyle\pm 0.09}$
                  & $1.29 {\scriptstyle\pm 0.10}$ & $6.84 {\scriptstyle\pm 0.30}$ & $2.60 {\scriptstyle\pm 0.06}$ \cr
        \hline
    \end{tabular}
    }
\end{table*}

As expected, ZSL approaches are extremely fast when disregarding the visual feature extraction phase. The low-dimension of both visual features and semantic representation of the classes grants a reduced number of computations, even when the algorithm used has polynomial time. In general, ZSL methods are capable of classifying visual features of an image in less than 1ms when using CPU, and in less than 25ms when using low-power devices. Regarding the comparison between projection methods and generative methods, it is not possible to conclude which strategy is faster. Instead, the difference lies in the type of models used, as the methods based on deep learning models have a higher processing time.

\begin{table}[h]
    \caption{Processing time of the visual feature extraction phase on CPU, Raspberry Pi 4B (R-PI 4B) and Jetson Nano. Execution time is presented in milliseconds with the format $avg \pm std$.\label{tab:feat_extract_times}}
    \centering
    \resizebox{0.475\textwidth}{!}{
    \begin{tabular}{|l|c|c|c|c|c|}
        \hline
        \multicolumn{1}{|l|}{} & \multicolumn{3}{|c|}{Processing Time} & \multicolumn{2}{|c|}{} \cr
        \hline
        \multicolumn{1}{|c|}{\textbf{Architecture}} & Desktop & R-PI 4B & Jetson Nano & Feat. Dim. & Size (MB) \cr
        \hline
        MobileNet & $\mathbf{25.57 {\scriptstyle \pm 3.17}}$ & $ 310.52 {\scriptstyle\pm 9.80}$ &  $\mathbf{29.58 {\scriptstyle\pm 3.14}}$ & $ 1024 $ & $ 16 $ \cr
        \hline
        MobileNetV2 & $ 27.59 {\scriptstyle\pm 3.38}$ & $\mathbf{297.63 {\scriptstyle\pm 8.54}}$ & $33.04 {\scriptstyle\pm 20.11}$ & $ 1280 $ & $ 14 $ \cr
        \hline
        InceptionV3 & $ 33.80 {\scriptstyle\pm 2.81}$ & $ 609.23 {\scriptstyle\pm 3.54}$ & $143.28 {\scriptstyle\pm 29.19}$ & $ 2048 $ & $ 92 $ \cr
        \hline
        ResNet50V2 & $ 38.07 {\scriptstyle\pm 3.13}$ & $ 887.86 {\scriptstyle\pm 5.27}$ & $160.87 {\scriptstyle\pm 1.78}$ & $ 2048 $ & $ 98 $ \cr
        \hline
        NASNetMobile & $ 39.67 {\scriptstyle\pm 2.35}$ & $ 370.10 {\scriptstyle\pm 5.79}$ & $111.81 {\scriptstyle\pm 20.40}$ & $ 1056 $ & $ 23 $ \cr
        \hline
        ResNet50 & $ 40.25 {\scriptstyle\pm 3.15}$ & $ 968.05 {\scriptstyle\pm 17.31}$ & $164.69 {\scriptstyle\pm 3.28}$ & $ 2048 $ & $ 98 $ \cr
        \hline
        Xception & $ 43.43 {\scriptstyle\pm 3.29}$ & $ 1081.18 {\scriptstyle\pm 11.07}$ & $155.75 {\scriptstyle\pm 23.21}$ & $ 2048 $ & $ 88 $ \cr
        \hline
        ResNet101V2 & $ 54.13 {\scriptstyle\pm 3.09}$ & $ 1655.37 {\scriptstyle\pm 15.72}$ & - & $ 2048 $ & $ 171 $ \cr
        \hline
        DenseNet201 & $ 54.79 {\scriptstyle\pm 2.86}$ & $ 1404.16 {\scriptstyle\pm 39.72}$ & - & $ 1920 $ & $ 80 $ \cr
        \hline
        ResNet101 & $ 57.46 {\scriptstyle\pm 2.99}$ & $ 1639.46 {\scriptstyle\pm 76.71}$ & - & $ 2048 $ & $ 171 $ \cr
        \hline
        VGG16 & $ 59.73 {\scriptstyle\pm 4.10}$ & $ 2046.51 {\scriptstyle\pm 15.03}$ & $221.75 {\scriptstyle\pm 16.96}$ & $ 512 $ & $ 528 $ \cr
        \hline
        VGG19 & $ 69.15 {\scriptstyle\pm 2.42}$ & $ 2557.60 {\scriptstyle\pm 6.12} $ & - & $ 512 $ & $ 549 $ \cr
        \hline
        EfficientNetB7 & $86.54 {\scriptstyle\pm 0.85}$ & $ 2436.62 {\scriptstyle\pm 106.28}$ & - & $2560$ & $256$ \cr
        \hline
        NASNetLarge & $ 95.67 {\scriptstyle\pm 2.98}$ & $ 2115.31 {\scriptstyle\pm 16.84}$ & - & $ 4032 $ & $ 343 $ \cr
        \hline
    \end{tabular}
    }
\end{table}

\subsection{Visual Feature Extraction: Inference Time\label{resuts_rq2}}

\begin{figure*}[h]
\resizebox{1.01\textwidth}{!}{%
\begin{tikzpicture}[font=\scriptsize]

        \node at (0.1,1.5) {\textbf{AWA2}};
        \node at (3.5+0.25\textwidth,1.5) {\textbf{CUB}};
        \node at (3.5+0.5\textwidth,1.5) {\textbf{APY}};
        \node at (3.5+0.75\textwidth,1.5) {\textbf{SUN}};
        
        \node[rotate=90] at (-2.35,0+0.25) {\textbf{ZSL}};
        \node[rotate=90] at (-2.35,-3+0.25) {\textbf{GZSL}};
        
        \node (awa2_zsl) at (0,0) {\includegraphics[width=0.245\textwidth]{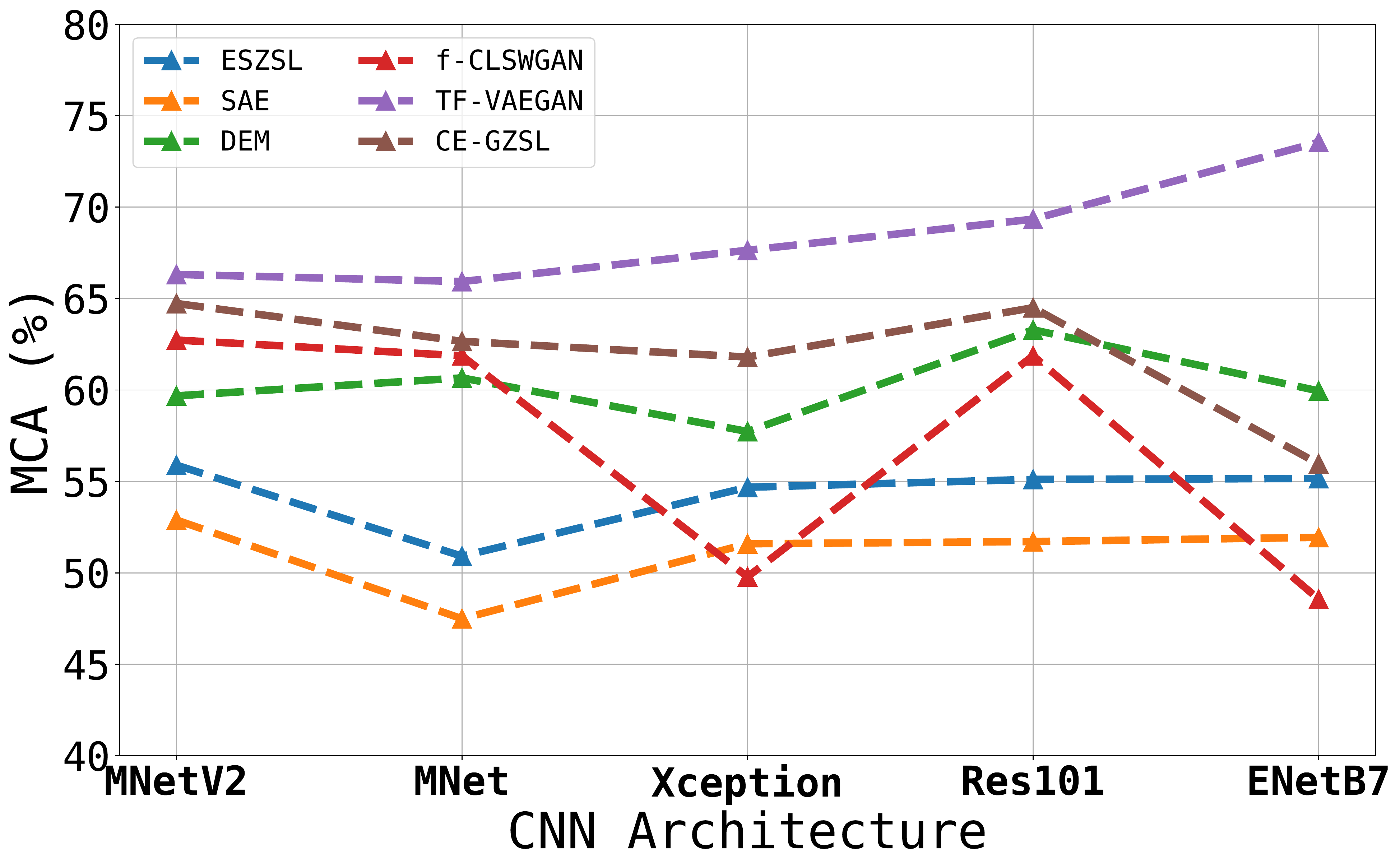}};
        \node (cub_zsl) at (0.25\textwidth,0) {\includegraphics[width=0.245\textwidth]{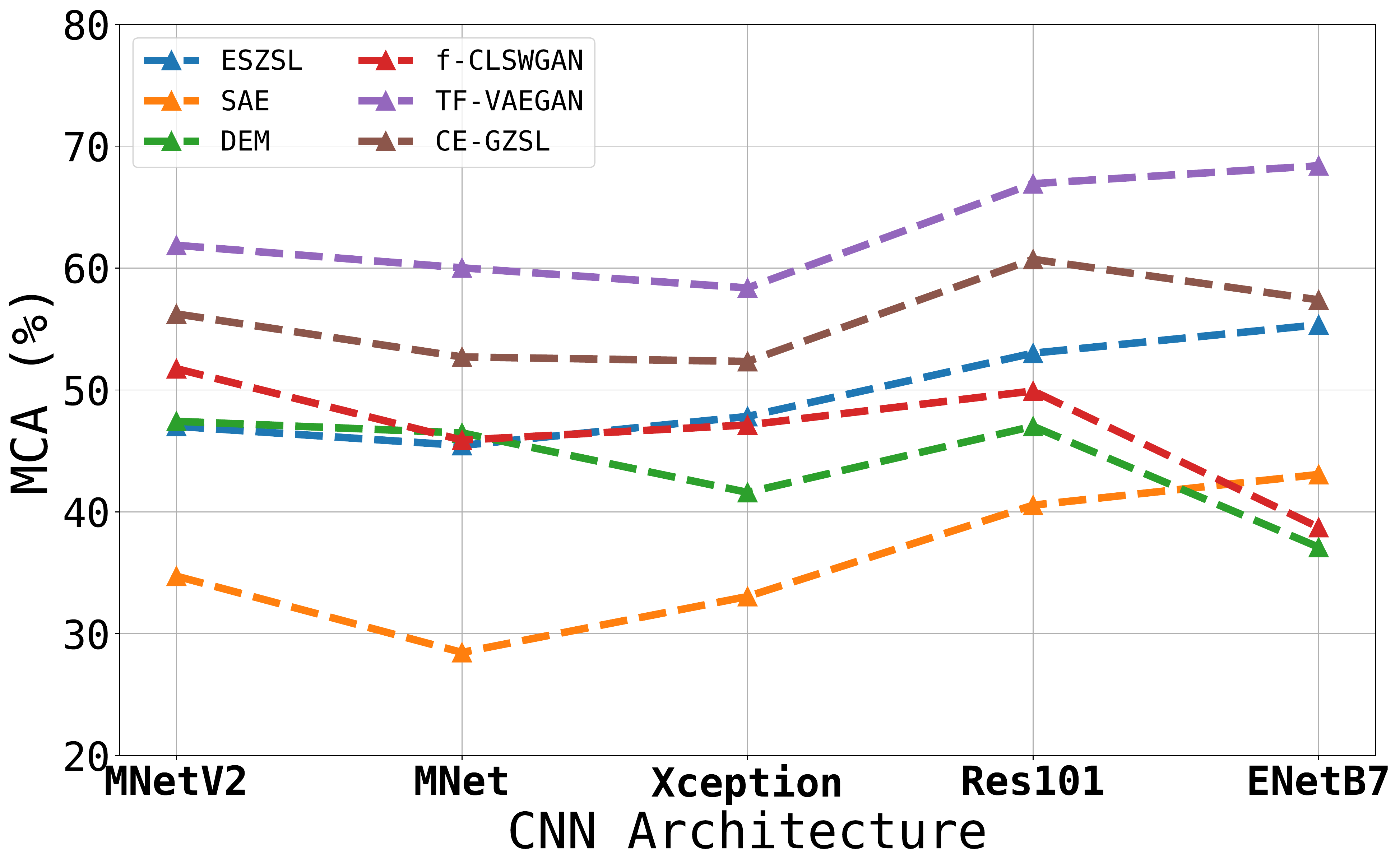}};
        \node (apy_zsl) at (0.5\textwidth,0) {\includegraphics[width=0.245\textwidth]{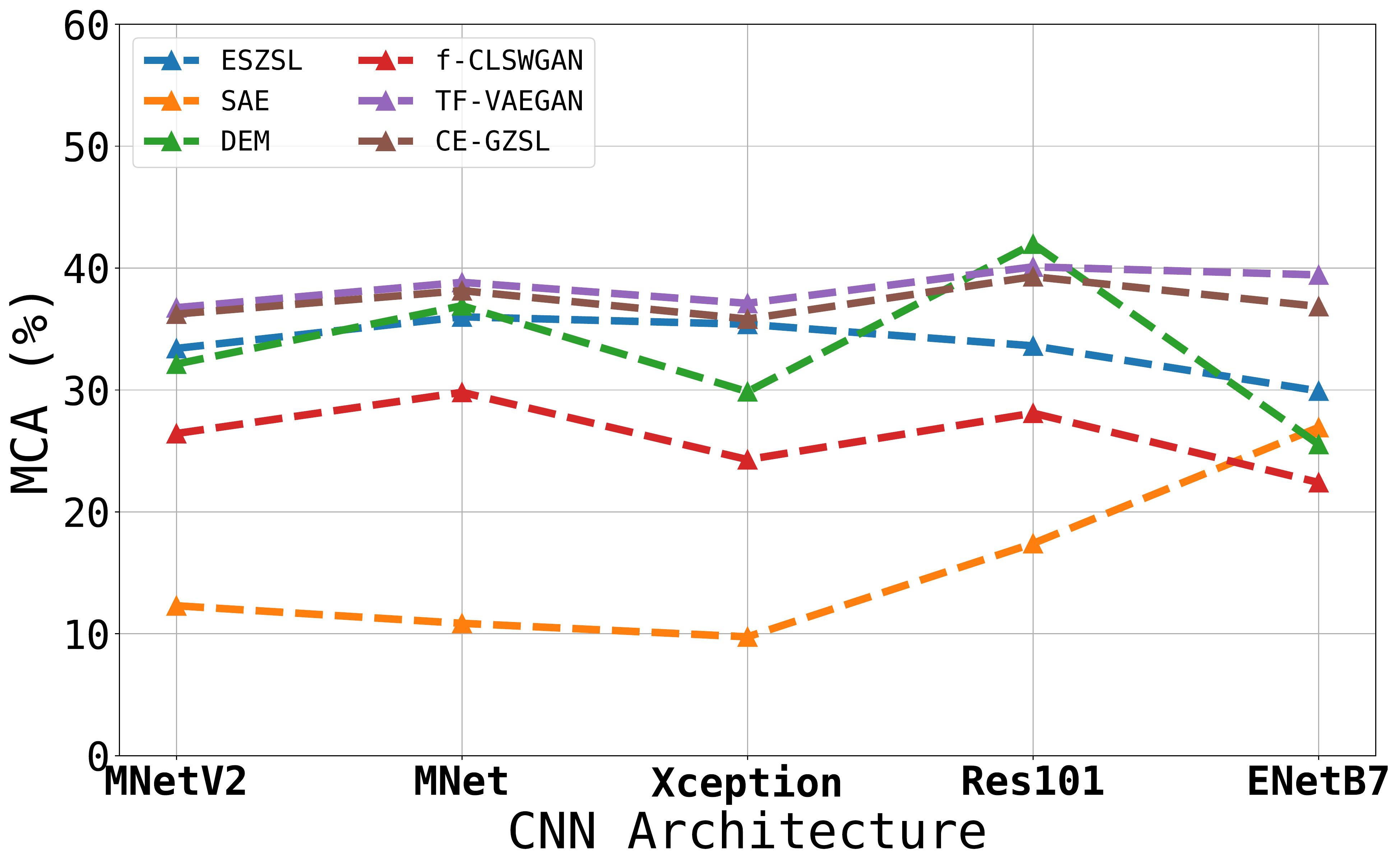}};
        \node (sun_zsl) at (0.75\textwidth,0) {\includegraphics[width=0.245\textwidth]{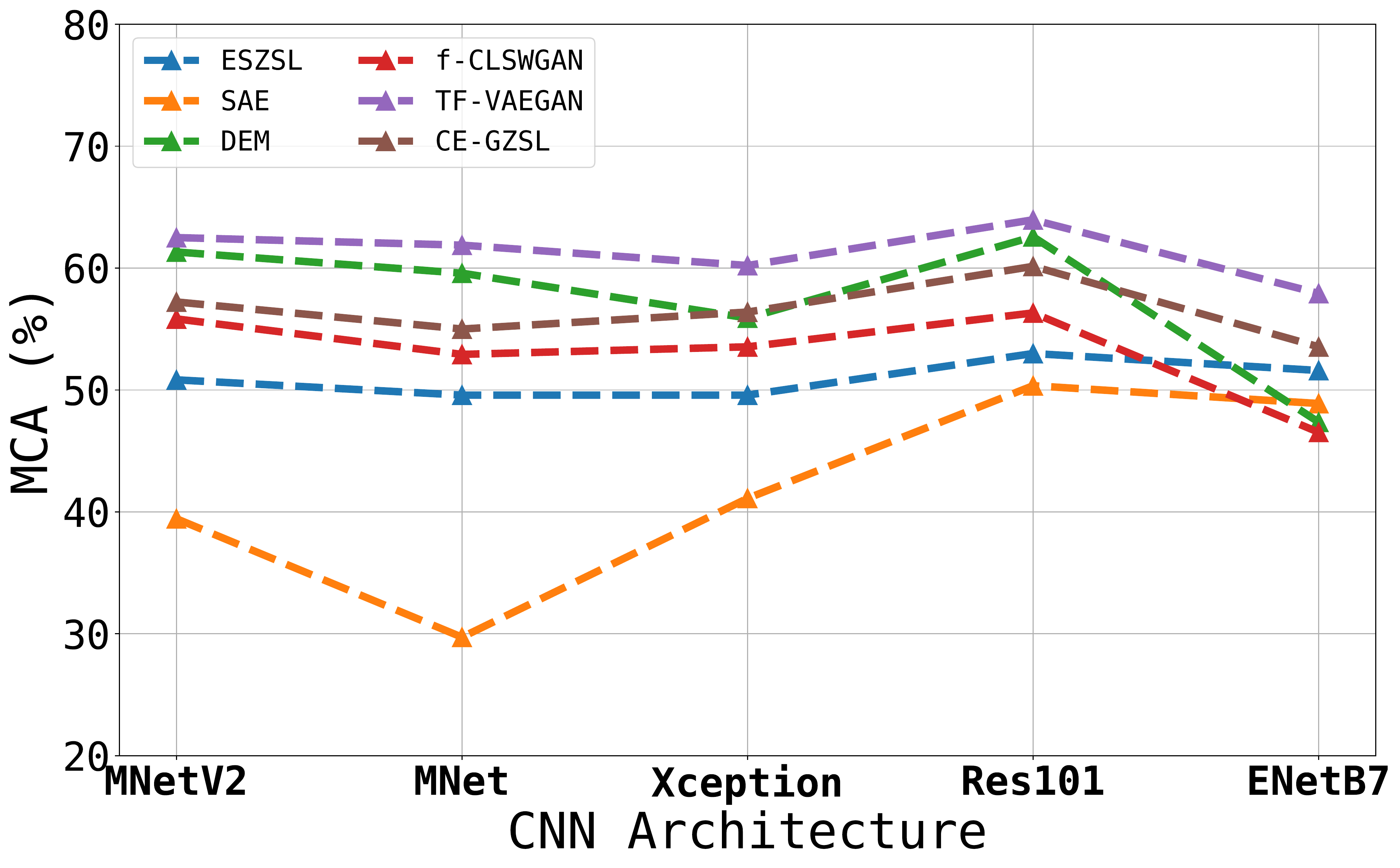}};
        
        \node (awa2_gzsl) at (0,-2.75) {\includegraphics[width=0.245\textwidth]{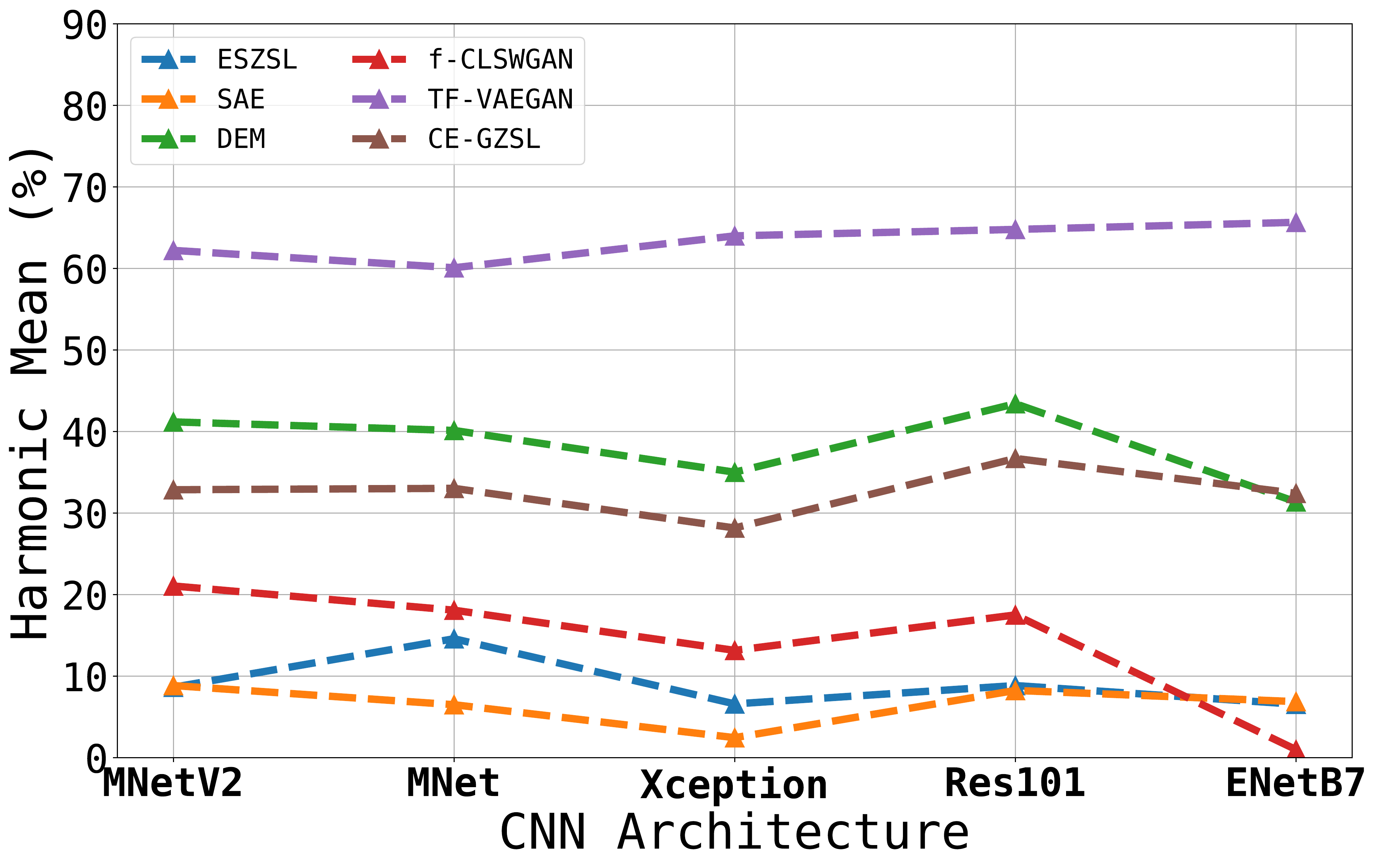}};
        \node (cub_gzsl) at (0.25\textwidth,-2.75) {\includegraphics[width=0.245\textwidth]{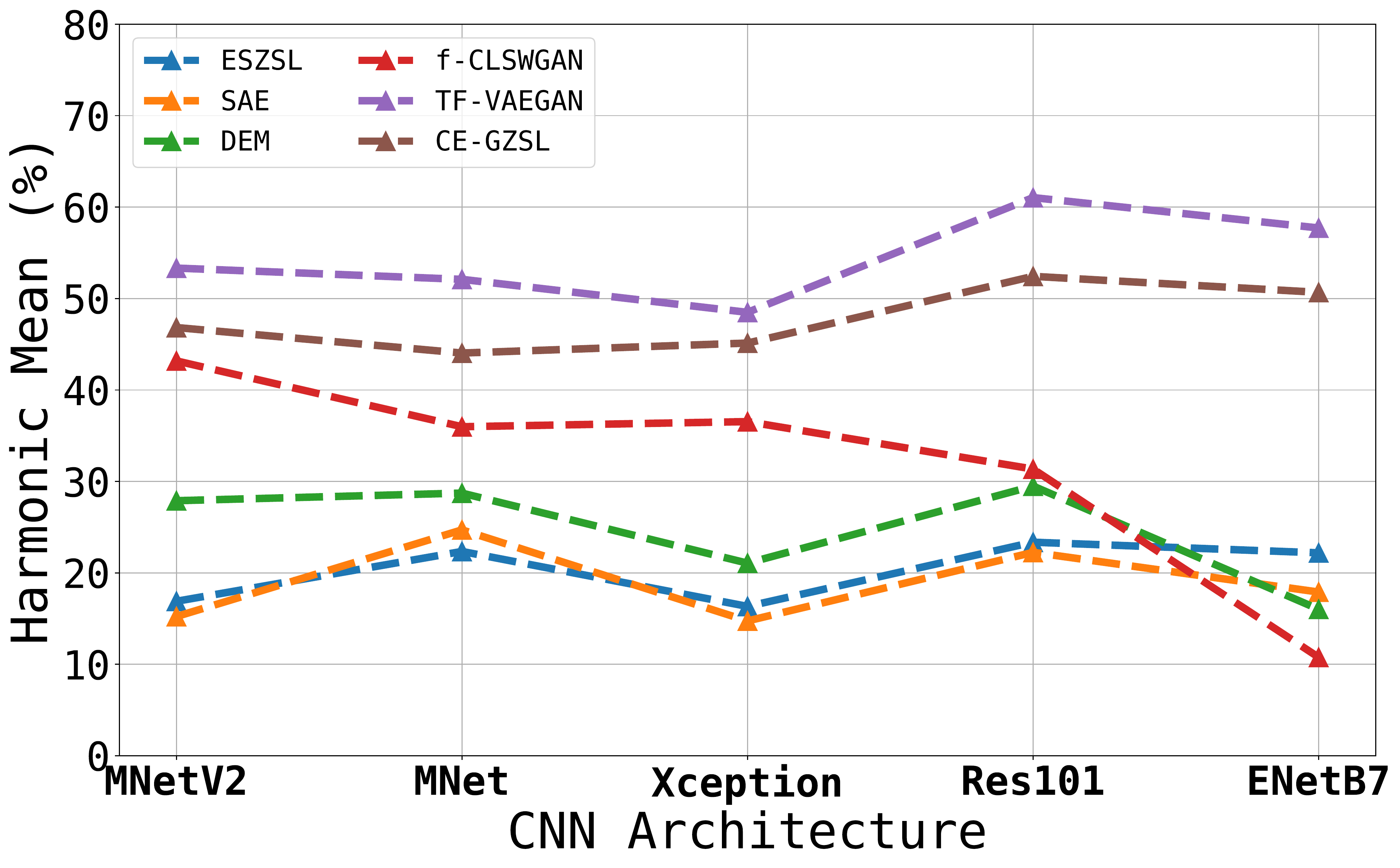}};
        \node (apy_gzsl) at (0.5\textwidth,-2.75) {\includegraphics[width=0.245\textwidth]{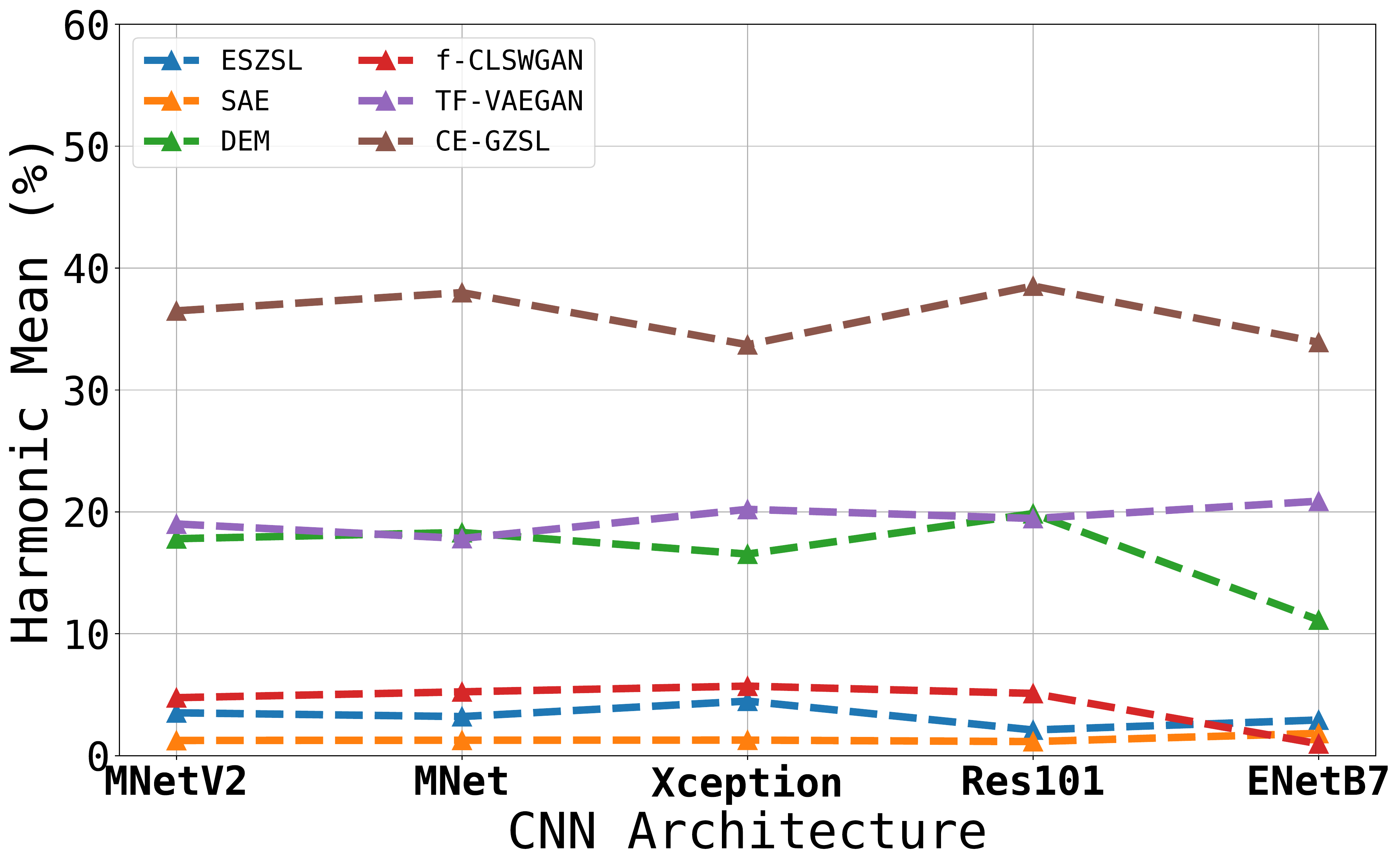}};
        \node (sun_gzsl) at (0.75\textwidth,-2.75) {\includegraphics[width=0.245\textwidth]{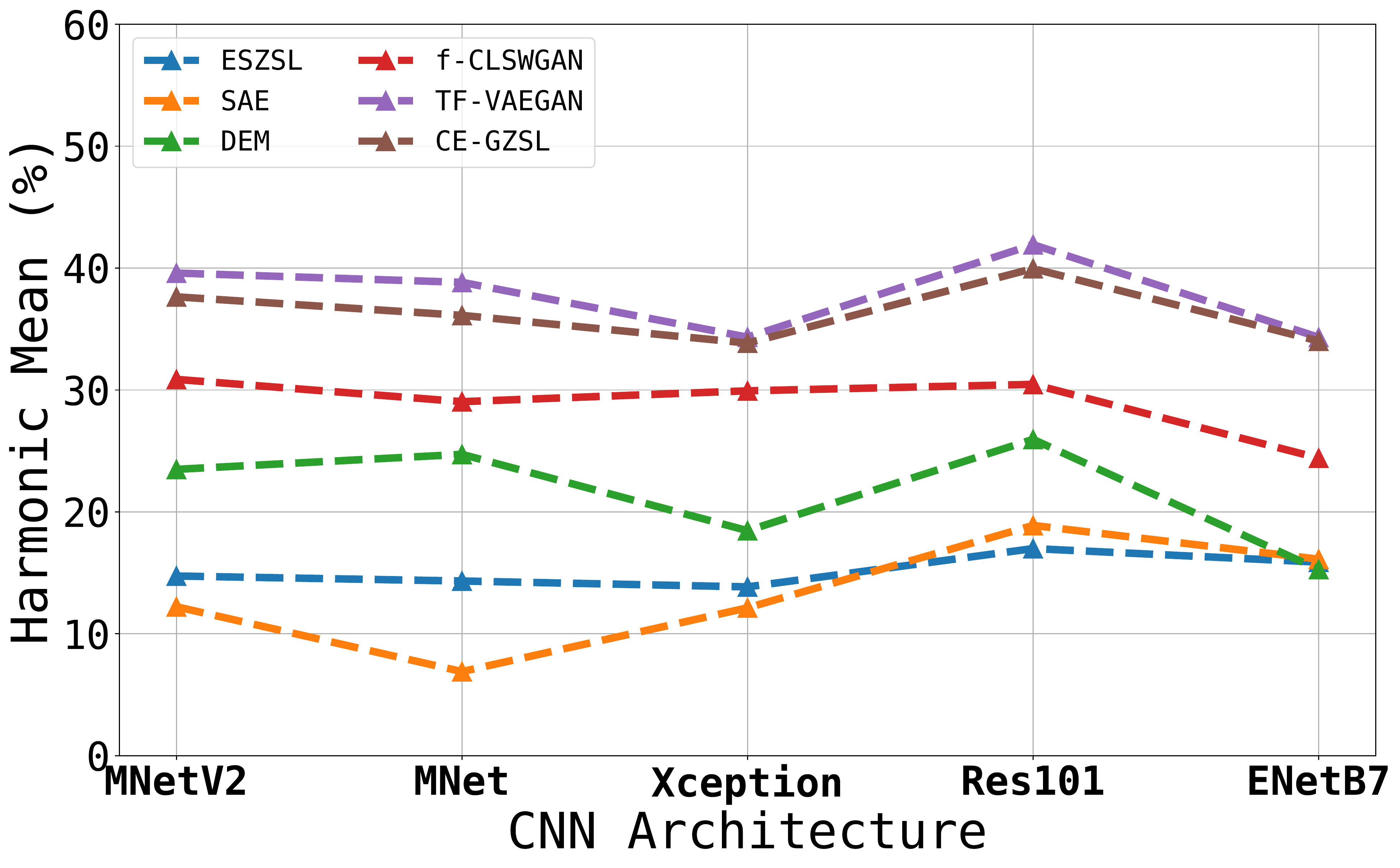}};
        
\end{tikzpicture}
}
\caption{\textbf{Performance of ZSL methods with respect to the CNN architecture used.} Six state-of-the-art ZSL approaches were trained using visual features obtained from CNN architectures with varying complexity (networks complexity increases from left to right in x-axis). The multi-way classification accuracy (MCA) of the evaluated approaches in four ZSL datasets evidences that lightweight architectures do not significantly reduce the performance of ZSL approaches.\label{fig:zsl_acc_speed}}
\end{figure*}
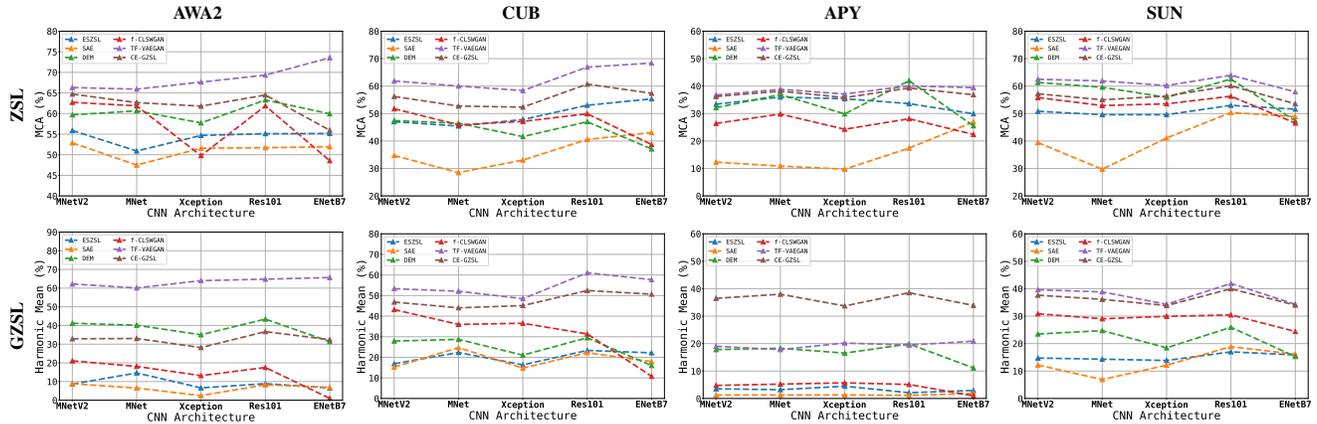

Considering that, in general, ZSL approaches are extremely fast during the inference stage (refer to section~\ref{resuts_rq1}), this section is devoted to the analysis of time consumed by the visual feature extraction phase, which is usually performed using CNNs. For this, we evaluate several CNN architectures with respect to the time consumed for obtaining the features of the top-layer pooling units, and using different hardware devices. The results are reported in Table~\ref{tab:feat_extract_times}. 

The results show that visual feature extraction is significantly slower when compared with the inference time of ZSL methods, being the bottleneck in overall inference stage of ZSL. Also, the results evidence that the consumed time varies largely over different architectures, and only lightweight models are capable of providing an acceptable running time in a low-power computational devices such as Raspberry Pi. However, these lightweight architectures are hardly used in ZSL applications, since the vast majority of the works adopt the ResNet101 architecture as the \textit{de facto} model for benchmarking the accuracy of ZSL methods. For this reason, it is particularly important to understand the impact of using other architectures in the accuracy of these methods, providing additional insights about their speed/accuracy trade-off regarding the architecture used for feature extraction.


\subsection{ZSL Accuracy: Impact of CNN Complexity}

Based on the observation that lightweight CNNs are capable of reducing significantly the time required to extract visual features, we argue that is particularly important to assess how the use of features obtained from lightweight models impacts the accuracy of ZSL methods. For this, the ZSL methods considered in this study are re-trained using the features extracted from networks with varying complexity. To ensure a fair evaluation, the hyper-parameters of each approach are adjusted during training using a validation set. Finally, each method is evaluated in the four datasets considered in this benchmark (AwA, CUB, APY and SUN) under the restricted and generalized (GZSL) settings. The results are depicted in Figure~\ref{fig:zsl_acc_speed}, and evidence that the features generated by lightweight architectures allow ZSL approaches to attain competitive results when compared with the \textit{de facto} ResNet101 architecture, typically used for assessing the accuracy of ZSL methods. Also, it can be observed that, in general, lines are nearly horizontal, meaning that the performance of ZSL methods does not consistently improve with the complexity of the architecture used for feature extraction. These results suggest that the inference phase of ZSL methods can be speed up without compromising the accuracy of ZSL.  

To provide a thorough evaluation of ZSL methods using different architectures, we also report in Table~\ref{tab:bench_mnetv2} the performance of each method using the standard metrics of ZSL, i.e.,  multi-way classification accuracy (MCA) in the restricted setting, and in the generalized setting the accuracy of the seen (S) and unseen (U) classes, as well as their harmonic mean (H).

\subsection{ZSL Speed/Accuracy Trade-off\label{sec:overall_inference_time}}

\begin{figure*}[t]
\resizebox{1.01\textwidth}{!}{%
\begin{tikzpicture}[font=\scriptsize]

        \node[anchor=west] at (1.9,1.5) {\textbf{AWA2}};
        \node[anchor=west] at (6.45,1.5) {\textbf{CUB}};
        \node[anchor=west] at (11,1.5) {\textbf{APY}};
        \node[anchor=west] at (15.55,1.5) {\textbf{SUN}};
        
        \node[rotate=90,anchor=west] at (-0.15,0) {\textbf{RPI4}};
        \node[rotate=90,anchor=west] at (-0.15,-3.5) {\textbf{Jetson Nano}};

        \node[anchor=west] (awa2_zsl) at (0.0041\textwidth,0) {\includegraphics[width=0.2459\textwidth]{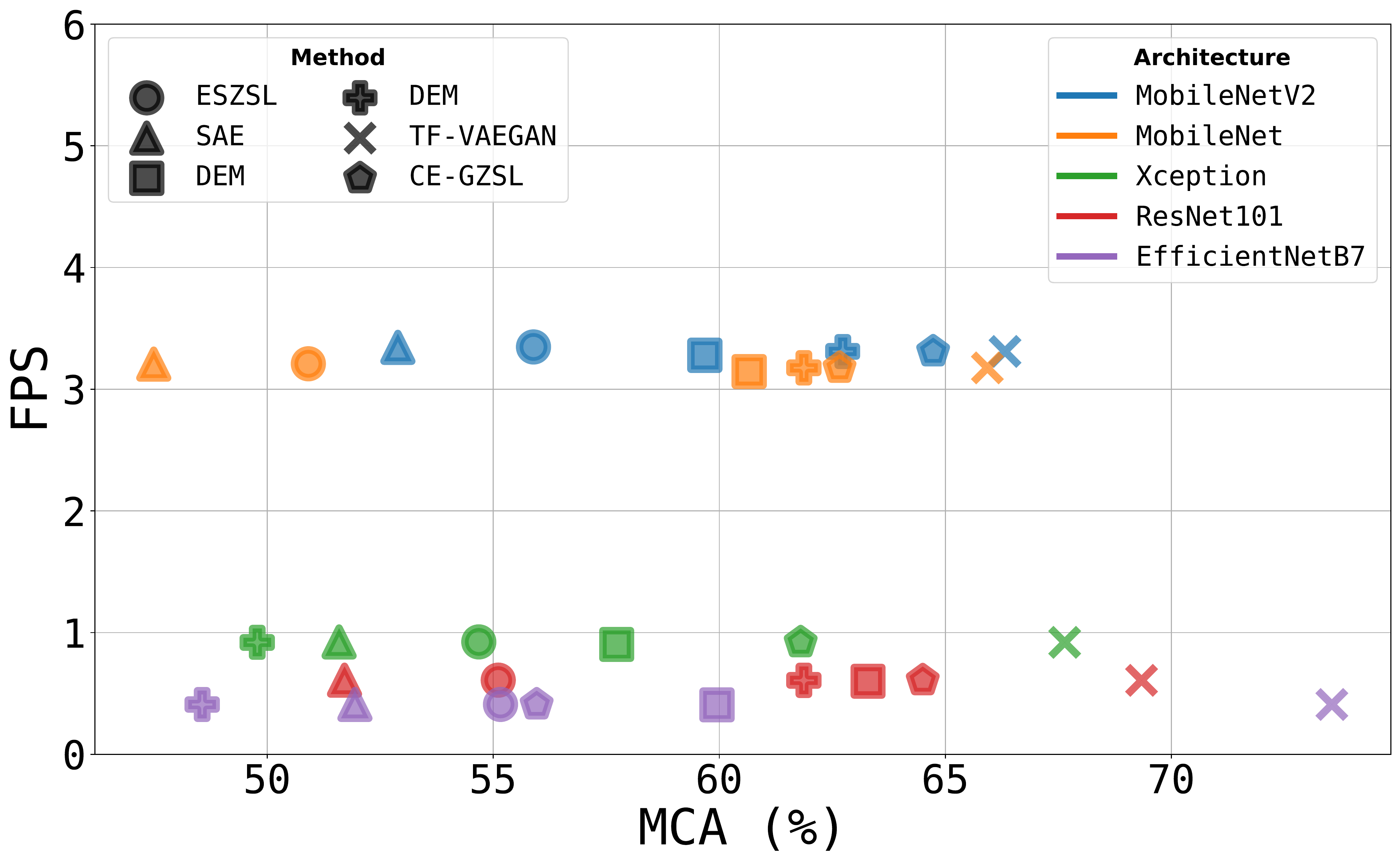}};
        \node[anchor=west] (cub_zsl) at (0.2591\textwidth,0) {\includegraphics[width=0.248\textwidth]{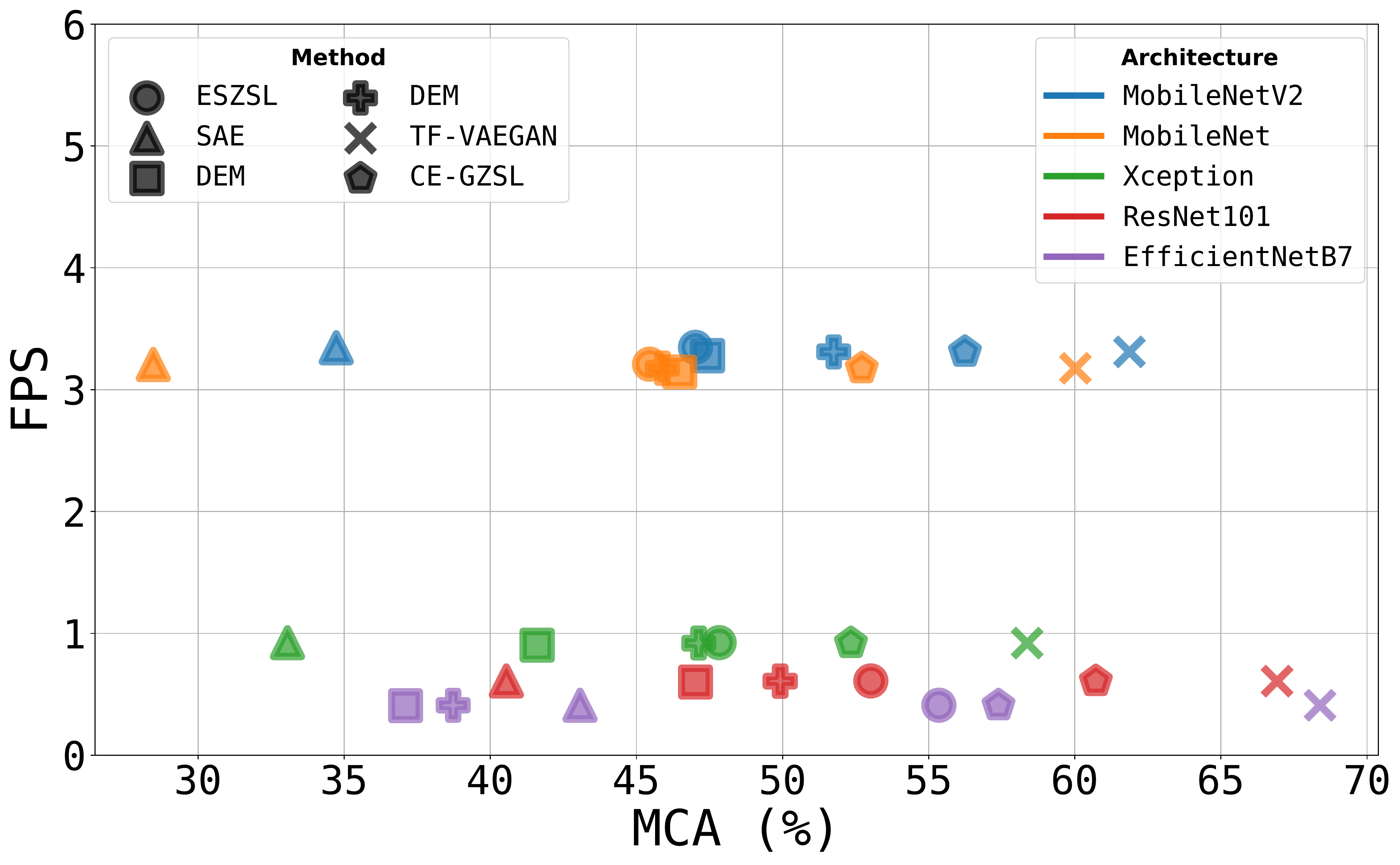}};
        \node[anchor=west] (apy_zsl) at (0.5141\textwidth,0) {\includegraphics[width=0.2445\textwidth]{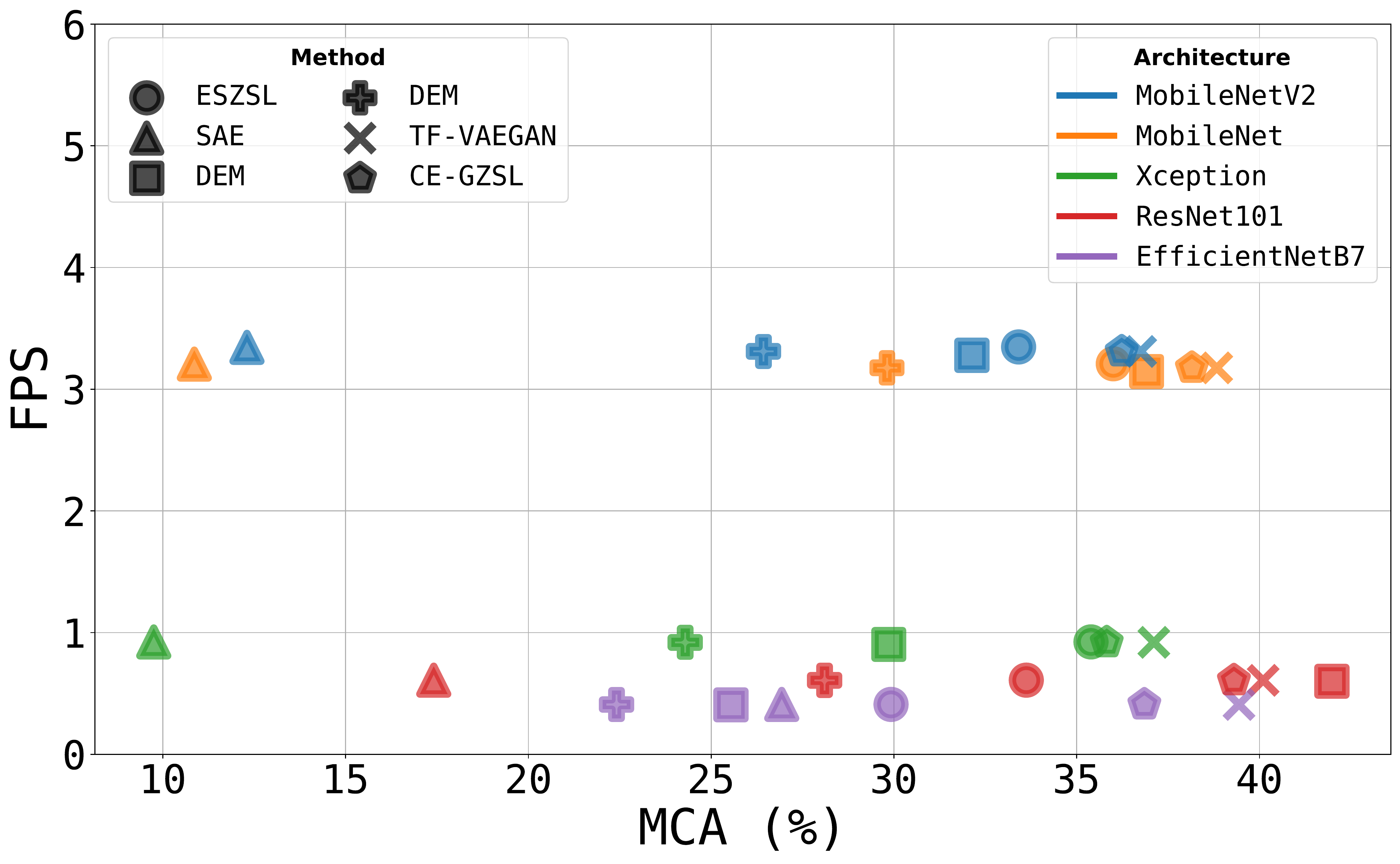}};
        \node[anchor=west] (sun_zsl) at (0.7691\textwidth,0) {\includegraphics[width=0.2457\textwidth]{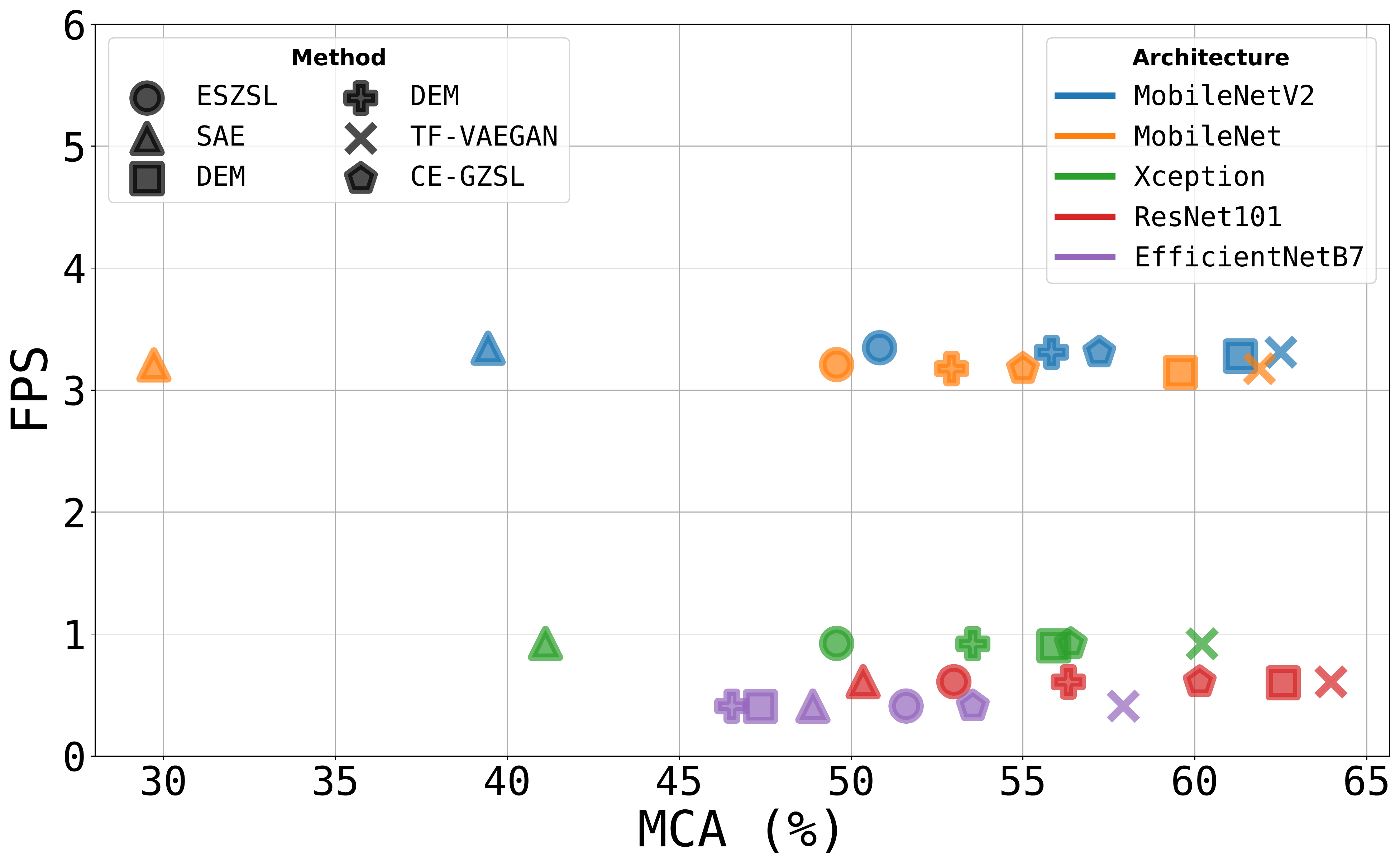}};
        
        \node[anchor=west] (awa2_gzsl) at (0,-2.75) {\includegraphics[width=0.25\textwidth]{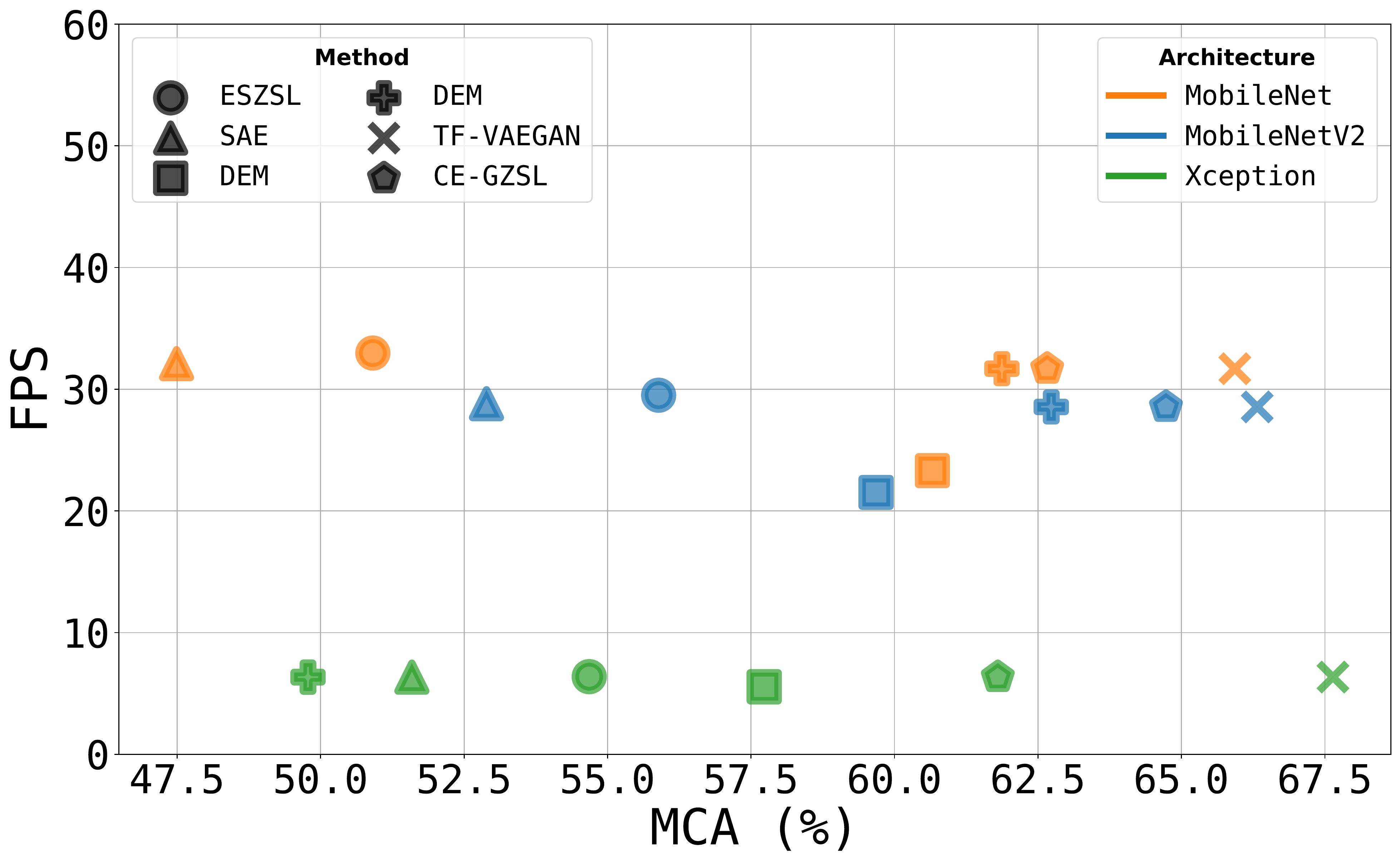}};
        \node[anchor=west] (cub_gzsl) at (0.255\textwidth,-2.75) {\includegraphics[width=0.25\textwidth]{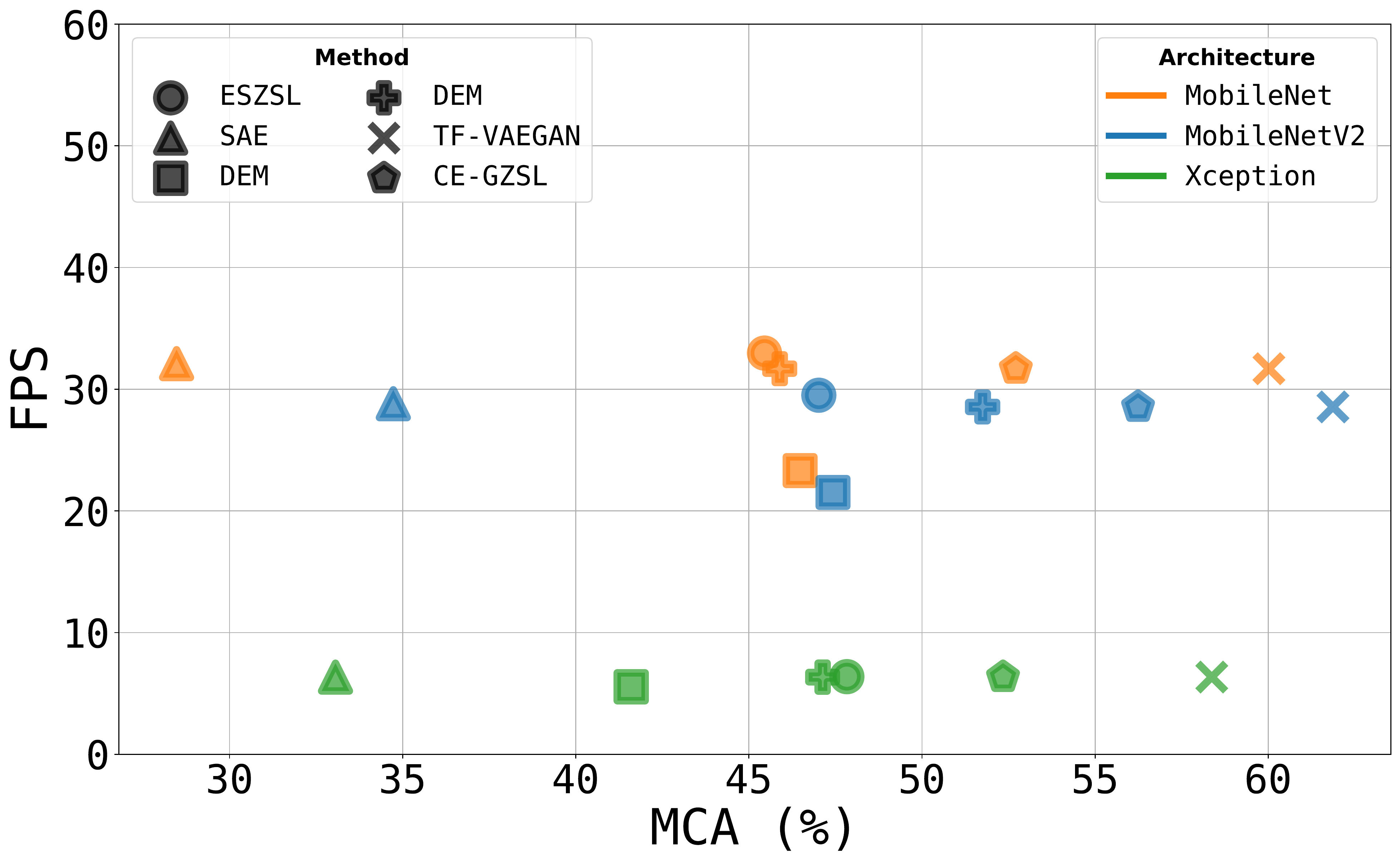}};
        \node[anchor=west] (apy_gzsl) at (0.51\textwidth,-2.75) {\includegraphics[width=0.25\textwidth]{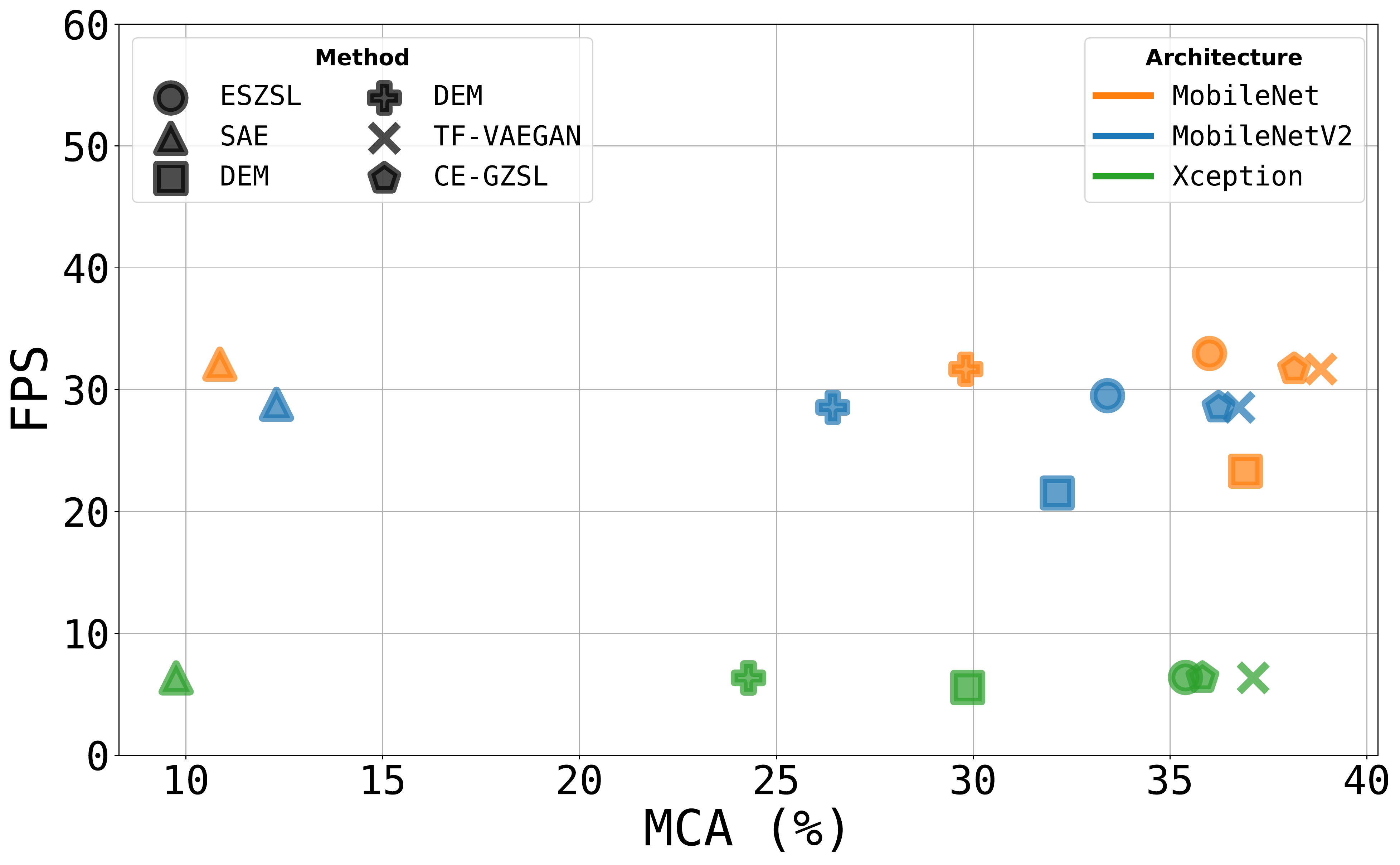}};
        \node[anchor=west] (sun_gzsl) at (0.765\textwidth,-2.75) {\includegraphics[width=0.25\textwidth]{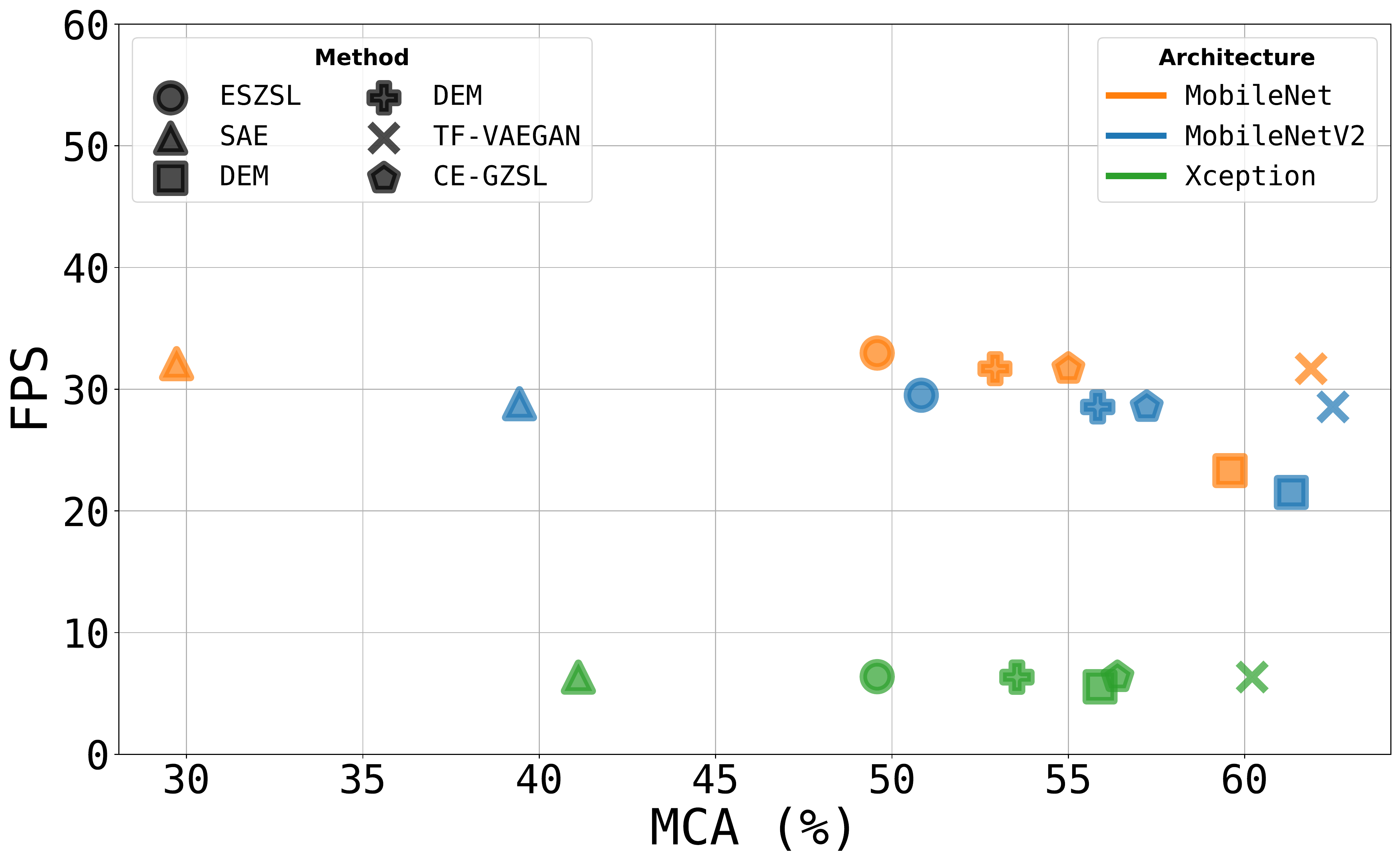}};
        
\end{tikzpicture}
}
\caption{\textbf{Accuracy/Speed trade-off of ZSL methods with respect to the CNN architecture used.} The accuracy and the overall inference time of ZSL methods is evaluated with respect to the architecture used for feature extraction. While lightweight networks are clearly faster than remaining architectures, it is interesting to note that the accuracy does not significantly decrease. Also, the FPS obtained using Jetson Nano evidence that ZSL methods can be seamlessly deployed in low-power computational devices if lightweight networks are used and the device is equipped with specialized hardware (refer to ~\ref{sec:overall_inference_time}).\label{fig:zsl_acc_speed_tradeoff}}
\end{figure*}
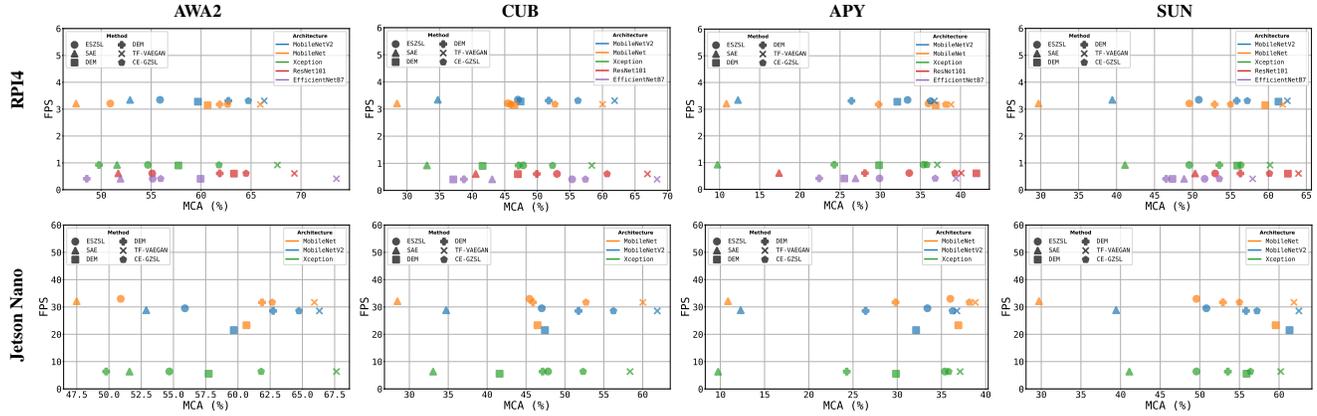

While section~\ref{resuts_rq1} analyzed the computational cost of the different phases of ZSL inference stage separately, it is particularly important to perceive how the processing time of the overall inference stage and the accuracy of ZSL methods is impacted by the use of different CNN architectures and the use of low-computational devices.\linebreak
Accordingly, we evaluate the performance of ZSL methods according to the MCA using four datasets commonly used in the field of ZSL (AwA, CUB, APY and SUN) under the restricted setting. The overall inference time of ZSL methods is measured according to the frames per second (FPS) processed when using low-power computational devices. Also, to ensure a fair and comprehensive evaluation of the performance of these devices, the complexity of CNN models is reduced by using different network optimization techniques, such as layer fusion, matrix normalization, and the reduction of floating point precision (FP16/INT8). These optimizations are carried out using the NVIDIA TensorRT, allowing the creation of models compatible with the integrated GPU of Jetson Nano. Figure~\ref{fig:zsl_acc_speed_tradeoff} depicts the results obtained, organized by dataset (columns) and the device used for inference (rows).\linebreak
As expected, lightweight networks significantly decrease the processing time of ZSL inference phase. However, the throughput of these networks does not exceed 4 FPS when using Raspberry Pi 4B, decreasing its applicability in real-world scenarios. In contrast, Jetson Nano is capable of delivering 30 FPS using lightweight networks, which can be explained by the optimizations performed using TensorRT and the the use of a integrated GPU. \linebreak
Regarding the comparison between lightweight architectures and the ResNet101, the top-1 accuracy decreases $4\%$ in average considering the four evaluated datasets, while the number of FPS increases from $0.6$ to $3.3$ in Raspberry Pi 4B, and $6.2$ to $30.6$ in Jetson Nano. 
This suggests that lightweight networks offer a compelling trade-off between inference time and accuracy, when compared to the \textit{de facto} architecure used for benchmarking ZSL accuracy.

\section{Conclusions}
This work introduces the first benchmark on zero-shot learning regarding the processing time in the inference stage. The two major phases of the inference stage (visual feature extraction and feature classification) of zero-shot learning approaches are analyzed with respect to the impact in the overall inference time. The obtained results confirm that visual feature extraction is the major bottleneck in the pipeline of ZSL approaches, justifying thus the need for studying the impact of using different CNN architectures in the processing time of this phase, as well as in the speed/accuracy trade-off of ZSL methods. Accordingly, we measure the accuracy and processing speed of different ZSL methods when using architectures of varying complexity. Also, the evaluation was carried out using different hardware devices, to assess the feasibility of deploying ZSL methods in low-power computational devices. The results suggest that the use of lightweight architectures does not significantly decreases the accuracy of ZSL methods, while reducing dramatically inference time. Moreover, the analysis to processing time of low-power devices shows that standard single-board computers are not sufficient to operate in real-world scenarios (maximum of 4 FPS), but the use of specialized hardware, such as integrated GPUs, in this type of low-power devices can significantly reduce the processing time of visual feature extraction, enabling to perform the inference at 30 FPS. \linebreak
Finally, we make publicly available a framework for simultaneously evaluating the accuracy and inference time of ZSL methods, which we hope to foster the advances on the topic of ZSL.

\section*{Acknowledgements}
The authors would like to thank the support provided by FCT strategic project NOVA LINCS (UIDB/04516/2020).

\begin{table*}[h]
    \centering
    \caption{Accuracy of ZSL methods evaluated under the restricted (ZSL) and generalized (GZSL) setting on the AWA2, CUB, SUN and APY datasets using different CNN architectures for the visual feature extraction.\label{tab:bench_mnetv2}}
    \resizebox{0.975\textwidth}{!}{
    \begin{tabular}{|l|c||c|ccc||c|ccc||c|ccc||c|ccc|}
        \hline
        \multicolumn{1}{|c|}{} & \multicolumn{1}{|c|}{} & \multicolumn{4}{|c|}{\textbf{AWA2}} & \multicolumn{4}{|c|}{\textbf{CUB}} & \multicolumn{4}{|c|}{\textbf{SUN}} & \multicolumn{4}{|c|}{\textbf{APY}}\cr
        \cline{3-18}
        \multicolumn{1}{|c|}{\textbf{Methods}} & \multicolumn{1}{|c|}{Architecture} & {ZSL} & \multicolumn{3}{|c|}{GZSL} & {ZSL} & \multicolumn{3}{|c|}{GZSL} & {ZSL} & \multicolumn{3}{|c|}{GZSL} & {ZSL} & \multicolumn{3}{|c|}{GZSL} \cr
        \cline{3-18} 
        \multicolumn{1}{|c|}{} & \multicolumn{1}{|c|}{} & MCA & U & S & H & MCA & U & S & H & MCA & U & S & H & MCA & U & S & H \cr
        \hline
        \multicolumn{1}{|c|}{} & ResNet101 & $55.11$ & $4.66$ & $87.07$ & $8.86$ & $53.02$ & $14.29$ & $63.73$ & $23.35$ & $52.99$ & $12.15$ & $28.22$ & $16.99$ & $33.62$ & $1.07$ & $72.24$ & $2.11$ \cr
        \multicolumn{1}{|c|}{} & MobileNet & $50.91$ & $8.04$ & $79.32$ & $14.60$ & $45.45$ & $14.22$ & $52.13$ & $22.34$ & $49.58$ & $10.49$ & $22.64$ & $14.33$ & $36.00$ & $1.65$ & $66.23$ & $3.21$ \cr
        ESZSL \cite{Romera_Paredes_ICML2015} & MobileNetV2 & $55.89$ & $4.56$ & $83.94$ & $8.65$ & $47.02$ & $9.96$ & $55.71$ & $16.89$ & $50.83$ & $10.76$ & $23.37$ & $14.74$ & $33.41$ & $1.81$ & $68.91$ & $3.52$ \cr
        \multicolumn{1}{|c|}{} & Xception & $54.68$ & $3.44$ & $86.67$ & $6.61$ & $47.83$ & $9.55$ & $56.23$ & $16.33$ & $49.58$ & $9.44$ & $25.93$ & $13.85$ & $35.39$ & $2.32$ & $69.75$ & $4.48$ \cr
        \multicolumn{1}{|c|}{} & EfficientNetB7 & $55.16$ & $3.42$ & $87.41$ & $6.57$ & $55.35$ & $13.45$ & $63.37$ & $22.19$ & $51.60$ & $11.25$ & $26.98$ & $15.88$ & $29.92$ & $1.50$ & $69.62$ & $2.93$ \cr
        \hline
        \multicolumn{1}{|c|}{} & ResNet101 & $51.71$ & $4.34$ & $85.39$ & $8.26$ & $40.55$ & $14.10$ & $52.55$ & $22.24$ & $50.35$ & $15.97$ & $23.10$ & $18.89$ & $17.41$ & $0.6$ & $18.02$ & $1.16$ \cr
        \multicolumn{1}{|c|}{} & MobileNet & $47.49$ & $3.40$ & $77.62$ & $6.51$ & $28.47$ & $15.87$ & $55.74$ & $24.70$ & $29.72$ & $5.69$ & $8.76$ & $6.90$ & $10.86$ & $0.69$ & $8.32$ & $1.27$\cr
        SAE \cite{Koridov_CVPR2017} & MobileNetV2 & $52.89$ & $4.66$ & $87.07$ & $8.86$ & $34.73$ & $9.21$ & $43.77$ & $15.22$ & $39.44$ & $9.93$ & $15.89$ & $12.22$ & $12.3$ & $0.71$ & $5.25$ & $1.25$ \cr
        \multicolumn{1}{|c|}{} & Xception & $51.59$ & $1.25$ & $87.64$ & $2.47$ & $33.06$ & $8.84$ & $44.32$ & $14.74$ & $41.11$ & $9.65$ & $16.32$ & $12.13$ & $9.75$ & $0.71$ & $6.68$ & $1.28$\cr
        \multicolumn{1}{|c|}{} & EfficientNetB7 & $51.94$ & $3.59$ & $85.70$ & $6.89$ & $43.07$ & $10.59$ & $57.61$ & $17.90$ & $48.89$ & $12.71$ & $22.05$ & $16.12$ & $26.93$ & $0.93$ & $53.70$ & $1.83$ \cr
        \hline
        \multicolumn{1}{|c|}{} & ResNet101 & $63.29$ & $29.21$ & $84.60$ & $43.42$ & $47.02$ & $21.56$ & $46.66$ & $29.49$ & $62.57$ & $20.0$ & $36.94$ & $25.95$ & $\mathbf{41.98}$ & $11.54$ & $71.28$ & $19.86$\cr
        \multicolumn{1}{|c|}{} & MobileNet & $60.66$ & $26.59$ & $81.87$ & $40.14$ & $46.48$ & $20.21$ & $49.61$ & $28.72$ & $59.58$ & $18.61$ & $36.78$ & $24.72$ & $36.91$ & $10.59$ & $67.87$ & $18.31$\cr
        DEM \cite{Zhang_CVPR2017} & MobileNetV2 & $59.68$ & $27.25$ & $84.37$ & $41.19$ & $47.43$ & $20.30$ & $44.59$ & $27.89$ & $61.32$ & $18.26$ & $32.91$ & $23.49$ & $32.13$ & $10.27$ & $66.71$ & $17.80$ \cr
        \multicolumn{1}{|c|}{} & Xception & $57.73$ & $21.98$ & $86.09$ & $35.02$ & $41.6$ & $15.78$ & $31.66$ & $21.06$ & $55.9$ & $13.89$ & $27.52$ & $18.46$ & $29.86$ & $9.61$ & $59.53$ & $16.54$\cr
        \multicolumn{1}{|c|}{} & EfficientNetB7 & $59.95$ & $19.14$ & $87.04$ & $31.38$ & $37.1$ & $11.49$ & $26.41$ & $16.02$ & $47.36$ & $12.15$ & $20.58$ & $15.28$ & $25.54$ & $6.37$ & $44.10$ & $11.14$ \cr
        \hline
        \multicolumn{1}{|c|}{} & ResNet101 & $61.87$ & $9.69$ & $\mathbf{90.89}$ & $17.52$ & $49.92$ & $20.93$ & $62.45$ & $31.35$ & $56.32$ & $28.47$ & $32.71$ & $30.46$ & $28.1$ & $2.64$ & $76.92$ & $5.11$\cr
        \multicolumn{1}{|c|}{} & MobileNet & $61.87$ & $10.07$ & $89.46$ & $18.10$ & $45.89$ & $27.13$ & $53.39$ & $35.98$ & $52.92$ & $33.13$ & $25.85$ & $29.04$ & $29.81$ & $2.71$ & $76.14$ & $5.24$\cr
        f-CLSWGAN \cite{Xian_CVPR2018} & MobileNetV2 & $62.73$ & $11.94$ & $89.73$ & $21.07$ & $51.75$ & $38.46$ & $49.22$ & $43.18$ & $55.83$ & $36.81$ & $26.59$ & $30.87$ & $26.43$ & $2.45$ & $\mathbf{78.61}$ & $4.76$ \cr
        \multicolumn{1}{|c|}{} & Xception & $49.78$ & $7.11$ & $89.96$ & $13.17$ & $47.13$ & $27.78$ & $53.39$ & $36.54$ & $53.54$ & $29.03$ & $30.89$ & $29.93$ & $24.29$ & $2.99$ & $64.41$ & $5.71$\cr
        \multicolumn{1}{|c|}{} & EfficientNetB7 & $48.56$ & $0.50$ & $90.12$ & $1.0$ & $38.73$ & $5.96$ & $55.39$ & $10.76$ & $46.53$ & $25.69$ & $23.26$ & $24.41$ & $22.41$ & $0.5$ & $75.82$ & $0.98$\cr
        \hline
        \multicolumn{1}{|c|}{} & ResNet101 & $69.34$ & $\mathbf{57.60}$ & $74.04$ & $64.79$ & $66.92$ & $\mathbf{57.91}$ & $64.52$ & $\mathbf{61.04}$ & $\mathbf{63.96}$ & $46.60$ & $\mathbf{38.10}$ & $\mathbf{41.92}$ & $40.1$ & $11.15$ & $76.54$ & $19.46$\cr
        \multicolumn{1}{|c|}{} & MobileNet & $65.93$ & $52.61$ & $70.06$ & $60.09$ & $60.02$ & $48.14$ & $56.76$ & $52.10$ & $61.88$ & $45.63$ & $33.80$ & $38.83$ & $38.83$ & $10.24$ & $68.57$ & $17.82$\cr
        TF-VAEGAN \cite{Narayan_2020} & MobileNetV2 & ${66.32}$ & $53.33$ & $74.65$ & ${62.22}$ & ${61.87}$ & $50.55$ & $56.43$ & ${53.33}$ & ${62.5}$ & $\mathbf{46.88}$ & $34.26$ & ${39.59}$ & ${36.75}$ & $10.99$ & $70.28$ & $19.0$ \cr
        \multicolumn{1}{|c|}{} & Xception & $67.64$ & $53.90$ & $78.76$ & $64.00$ & $58.37$ & $44.52$ & $53.24$ & $48.49$ & $60.21$ & $39.51$ & $30.35$ & $34.33$ & $37.11$ & $11.81$ & $72.20$ & $20.21$\cr
        \multicolumn{1}{|c|}{} & EfficientNetB7 & $\mathbf{73.55}$ & $54.92$ & $81.62$ & $\mathbf{65.66}$ & $\mathbf{68.39}$ & $51.43$ & $\mathbf{65.77}$ & $57.72$ & $57.92$ & $36.04$ & $32.75$ & $34.32$ & $39.44$ & $12.05$ & $78.25$ & $20.88$\cr
        \hline
        \multicolumn{1}{|c|}{} & ResNet101 & $64.5$ & $23.03$ & $90.46$ & $36.71$ & $60.72$ & $56.75$ & $48.72$ & $52.43$ & $60.14$ & $45.14$ & $35.85$ & $39.96$ & $39.3$ & $\mathbf{31.07}$ & $50.69$ & $\mathbf{38.53}$\cr
        \multicolumn{1}{|c|}{} & MobileNet & $62.66$ & $20.39$ & $86.98$ & $33.04$ & $52.71$ & $39.89$ & $49.17$ & $44.04$ & $55.0$ & $40.14$ & $32.83$ & $36.12$ & $38.15$ & $29.40$ & $53.67$ & $37.99$\cr
        CE-GZSL \cite{Han_2021} & MobileNetV2 & $64.73$ & $20.18$ & $88.56$ & $32.87$ & $56.24$ & $48.37$ & $45.38$ & $46.83$ & $57.22$ & $44.10$ & $32.83$ & $37.64$ & $36.23$ & $30.82$ & $44.72$ & $36.49$\cr
        \multicolumn{1}{|c|}{} & Xception & $61.8$ & $16.71$ & $89.80$ & $28.18$ & $52.34$ & $39.22$ & $53.14$ & $45.13$ & $56.39$ & $38.33$ & $30.31$ & $33.85$ & $35.82$ & $11.81$ & $72.20$ & $20.21$\cr
        \multicolumn{1}{|c|}{} & EfficientNetB7 & $55.96$ & $19.76$ & $90.35$ & $32.43$ & $57.39$ & $50.24$ & $51.14$ & $50.68$ & $53.54$ & $37.15$ & $31.36$ & $34.01$ & $36.85$ & $25.87$ & $49.19$ & $33.91$\cr
        \hline
    \end{tabular}
    }
\end{table*}

{\small
\bibliographystyle{ieeetr}
}

\end{document}